\documentclass{article}
\pdfoutput=1
    \PassOptionsToPackage{numbers, compress}{natbib}
    \PassOptionsToPackage{dvipsnames}{xcolor}
    \usepackage[preprint]{neurips_2020}

\usepackage[utf8]{inputenc} % allow utf-8 input
\usepackage[T1]{fontenc}    % use 8-bit T1 fonts
\usepackage{hyperref}       % hyperlinks
\usepackage{url}            % simple URL typesetting
\usepackage{booktabs}       % professional-quality tables
\usepackage{amsfonts}       % blackboard math symbols
\usepackage{nicefrac}       % compact symbols for 1/2, etc.
\usepackage{microtype}      % microtypography

\usepackage{amsmath,amsfonts,bm}

\def\eqref#1{equation~\ref{#1}}
\def\1{\bm{1}}

\DeclareMathAlphabet{\mathsfit}{\encodingdefault}{\sfdefault}{m}{sl}
\SetMathAlphabet{\mathsfit}{bold}{\encodingdefault}{\sfdefault}{bx}{n}

\DeclareMathOperator*{\argmax}{arg\,max}
\DeclareMathOperator*{\argmin}{arg\,min}

\usepackage{tikz}
\usepackage{graphicx}
\graphicspath{ {experiments/} }
\usepackage{float}
\usepackage[margin=1cm,font=small,labelfont=bf]{caption}
\usepackage{wrapfig}
\usepackage{subcaption}
\usepackage{stfloats}
\usepackage{mathtools}
\usepackage{seqsplit}
\usepackage{longtable}

\usepackage{enumitem}

\newcommand\Tstrut{\rule{0pt}{2.6ex}}         % = `top' strut
\newcommand\Bstrut{\rule[-0.9ex]{0pt}{0pt}}   % = `bottom' strut

\title{{\LARGE Wat zei je?}\\
Detecting Out-of-Distribution Translations with Variational Transformers}

\author{
Tim Z. Xiao \\
\texttt{\small tim.z.xiao@outlook.com}
\And
Aidan N. Gomez \\
\texttt{\small aidan.gomez@cs.ox.ac.uk}
\And
Yarin Gal \\
\texttt{\small yarin@cs.ox.ac.uk}
\AND
\textnormal{Oxford Applied and Theoretical Machine Learning Group} \\
Department of Computer Science \\
University of Oxford
}

\begin{document}

\maketitle

\begin{abstract}

We detect out-of-training-distribution sentences in Neural Machine Translation using the Bayesian Deep Learning equivalent of Transformer models. For this we develop a new measure of uncertainty designed specifically for long sequences of discrete random variables---i.e.\ words in the output sentence. 
% This measure is able to convey epistemic uncertainty akin to the Mutual Information (MI), which is used in the case of single discrete random variables such as in classification. 
Our new measure of uncertainty solves a major intractability in  the naive application of existing approaches on long sentences.
We use our new measure on a Transformer model trained with dropout approximate inference.
On the task of German-English translation using WMT13 and Europarl, we show that with dropout uncertainty our measure is able to identify when Dutch source sentences, sentences which use the same word types as German, are given to the model instead of German. 
% We finish with a comprehensive stufy of the new methodology and discuss potential .
\end{abstract}

\section{Introduction}

Statistical Machine Translation (SMT, \citep{brown1993mathematics, och2003minimum}), built on top of probabilistic modelling foundations such as the IBM alignment models \citep{vogel1996hmm, brown1993mathematics, gal-blunsom-2013-systematic}, has largely been replaced in recent years following the emergence of Neural Machine Translation approaches (NMT, \citep{kalchbrenner-blunsom-2013-recurrent-continuous, bahdanau2014neural, luong2015effective, vaswani2017attention}). 
This change has brought with it huge performance gains to the field \citep{sennrich2016edinburgh}, but at the same time we have lost many desirable properties of these models. Statistical probabilistic models can inform us when they are guessing at random on inputs they never saw before \citep{ghahramani2015probabilistic}. This information can be used, for example, to detect out-of-training-distribution examples for selective classification by referring uncertain inputs to an expert for annotation \citep{leibig2017leveraging}, or for a human-in-the-loop approach to reduce data labelling costs \citep{gal2017deep, walmsley2019galaxy, kirsch2019batchbald}. 

With new tools in machine learning we can now incorporate such probabilistic foundations into deep learning NLP models without sacrificing performance. This field, known as Bayesian Deep Learning (BDL, \citep{neal2012bayesian, gal2016uncertainty}), is concerned with the development of scalable tools which capture \textit{epistemic} uncertainty---the model's notion of ``I don't know'', a measure of a model's lack of knowledge e.g.\ due to lack of training data, or when an input is given to the model which is very dissimilar to what the model has seen before. Such BDL tools have been used extensively in the Computer Vision literature \citep{kendall2017uncertainties, litjens2017survey}, and have been demonstrated to be of practical use for applications including medical imaging \citep{litjens2017survey, nair2020exploring}, robotics \citep{gal2016improving, chua2018deep}, and astronomy \citep{hon2018detecting, soboczenski2018bayesian, hezaveh2017fast}. 

In this paper we extend these tools, often used for vision tasks, to the language domain. We demonstrate how these tools can be used effectively on the task of selective classification in NMT by identifying source sentences the translation model has never seen before, and referring such source sentences to an expert for translation. We demonstrate this with state-of-the-art Transformer models, and show how model performance increases when rejecting sentences the model is uncertain about---i.e.\ the model's measure of epistemic uncertainty correlates with mistranslations. 

For this we develop several measures of epistemic uncertainty for applications in natural language processing, concentrating on the task of machine translation (\S\ref{sect:measures}). We compare these measures both with standard deterministic Transformer models, as well as with Variational Transformers, a new approach we introduce to capture epistemic uncertainty in sequence models using dropout approximate inference \citep{gal2016uncertainty}. We give an extensive analysis of the methodology, and compare the different approaches quantitatively in the out-of-training-distribution settings (\S\ref{sect:exps}), which shows our purposed uncertainty estimate BLEUVar works well for measuring the epistemic uncertainty for machine translation. 
We also analyse the performance of BLEUVar qualitatively from both the influence of sentence length and from the linguistic perspective.
We finish with a discussion in potential use cases for the new methodogy proposed.
%We show how training set size affects the proposed measures, assess model calibration, and detect out-of-training-distributions inputs as well as distribution shifts. 
% We finish by surveying related work to our own (\S\ref{sect:related}), and discuss potential use cases for the new methodology proposed (\S\ref{sect:concs}).

The closest NLP task to the above problem definition is the quality estimation (QE) task in Machine Translation \citep{specia2010machine, blatz2004confidence}, which tries to solve a similar problem by predicting the quality of a translation with a score called Human Translation Error Rate (HTER, \citep{hter}). This is done by training a surrogate QE model on source sentences and their corresponding machine-generated translations in a specific domain, with the target of the surrogate to predict the the percentage of edits needed to be fixed. While many methods have been shown to successfully solve the task of estimating the quality of translations \citep{kim2017predictor, martins2016unbabel, martins2017pushing, kreutzer2015quality}, by definition QE crucially relies on examples of mistranslations to train the surrogate. The assumption that such training data is available is often violated in practice though (e.g.\ in active learning), thus existing approaches in QE research cannot generally be used to detect out-of-training-distribution examples (see Appendix \ref{sect:related} for detail discussion about the differences between QE and our task, as well as other related work that is similar to ours but not solving the same problem).

\section{Background: Uncertainty in Deep Learning}

% \tim{Approximate Inference}

For most machine learning models, the optimisation objectives give us a point estimate of the model parameters, which maximise the likelihood of the model generating the training data (i.e. $p({Y}|{X},{\omega} =  {\omega}^\ast)$, $ {\omega}^\ast \in  {\Omega}$ s.t. $ {\Omega}$ is the set of all possible model parameters, $ {Y}, {X}$ are the training data). Such point estimate will provide us with a very good prediction when the test data follows the same distribution as the training data distribution. Given a new input $ {x}^*$ at test time, the model prediction for the corresponding $ {y}^*$ is
\begin{equation}
    \hat{ {y}}^* = \argmax_{ {y}^*} p( {y}^*| {x}^\ast, {\omega}^\ast).
\end{equation}
However, we cannot expect the model to perform well on out-of-distribution (OOD) data which it never saw before. Instead, we would wish for the model to indicate its uncertainty towards such inputs. We could use $p(  \hat{ {y}}^*| {x}^\ast, {\omega}^\ast)$ as an estimate for model uncertainty, but as we show below, it would not be a well calibrated one. 
% $ {\omega}^\ast$ being a point estimate implies that we ignore the variance of $ {\omega}^\ast$ in the first place. 
It might be the case that many $ {\omega}$ might give equally good predictions on the train set, but might widely disagree with their predictions on OOD data. In fact, $ {\omega}^\ast$ might give arbitrary predictions on OOD training data which is very dissimilar to previously observed inputs. Thus, a high score does not distinguish whether $ {x}^\ast$ is OOD or not, and whether we should trust the model's prediction.

\subsection{Bayesian Inference}

Bayesian probabilistic models capture the notion of uncertainty explicitly. Rather than considering a single point estimate $ {\omega}^\ast$, Bayesian models aim to derive the entire distribution of $ {\omega}$ from the training data. The resulting distribution is called \textbf{posterior} distribution
\begin{equation}
    p( {\omega} |  {X,Y}) = \frac{p( {Y}| {X}, {\omega})p( {\omega})}{p( {Y}| {X})}.
\end{equation}
At test time, we can make prediction about the corresponding $ {y}^*$ by integrating out all possible $ {\omega}$
\begin{equation}
    p( {y}^* |  {x}^*, {X,Y}) = \int p( {y}^* |  {x}^*, {\omega}) p( {\omega}| {X,Y})\mathrm{d} {\omega}.
\end{equation}
Using the variance of the predictive distribution $p( {y}^* |  {x}^*, {X,Y})$ as the uncertainty measure would have taken into account the variance of $ {\omega}$. Hence, an uncertainty measure based on this quantity could be a strong indicator for ${x}^*$ being OOD.

\subsection{Approximate Inference}

The difficulty for doing Bayesian inference comes from the intractability of calculating the evidence% integral
\begin{equation}
    p( {Y}| {X}) = \int p( {Y}| {X}, {\omega}) p( {\omega}) \mathrm{d} {\omega}.
\end{equation}
There might be closed form solution for a simple model. But for most interesting problems, it is too difficult to compute an exact solution. Therefore, approximations are often used for such inference problems.

Variational inference (VI) is a pragmatic popular method for doing approximate inference \citep{jordan1999}. The method involves defining an approximating distribution $q_{\theta}( {\omega})$, and trying to find the parameter $\theta$ for that minimises the Kullback-Leibler (KL) divergence \citep{kullback1951} between $q_{\theta}( {\omega})$ and the posterior
\begin{align}
    \mathrm{KL}(q_{\theta}( {\omega}) \| p( {\omega} |  {X,Y})) = \int q_{\theta}( {\omega}) \mathrm{log} \frac{q_{\theta}( {\omega})}{p( {\omega} |  {X,Y})} \mathrm{d} {\omega}.
\end{align}
The resulting $\theta^*$ gives us the closest approximation for the posterior in the family of distribution $q_{\theta}( {\omega})$. However, calculating the KL divergence here is also intractable as we still have to do the integration for the evidence in posterior.

Fortunately, a tractable and equivalent objective for minimising the KL divergence is to maximise the evidence lower bound (ELBO, \citep{blei2017}) of $q_{\theta}( {\omega})$
\begin{align}
    \mathrm{ELBO}(q_{\theta}) = \int q_{\theta}({\omega}) \mathrm{log} \, p(  {Y}|  {X},  {\omega}) \mathrm{d}   {\omega} - \mathrm{KL}(q_{\theta}(  {\omega}) \| p(  {\omega})).
\end{align}
This is tractable as we know all the distributions (both $q_{\theta}({\omega})$, $p({Y}|{X},{\omega})$ and $p({\omega})$ are defined by the model or by our assumptions).

\subsection{Bayesian Inference in Deep Learning}

Almost all deep models treat the units in a deep neural network as deterministic functions. To adapt Bayesian methods in deep learning, we need to first turn our model into a probabilistic model. It can be done by modelling the weights in each unit of the network as samples from probability distributions. Such networks are called Bayesian neural networks \citep{neal2012bayesian}. One major challenge with Bayesian neural networks is that the integration in ELBO becomes intractable when we have more than one hidden layer \citep{gal2016uncertainty}.

Many works have tried tackling this problem. One practical method proposed by \citet{gal2016uncertainty} is MC Dropout. \citet{gal2016uncertainty} showed that optimising any neural networks with dropout can be viewed as an approximate inference in a probabilistic model (when dropout $p$ is tuned correctly), which implies that a trained neural network with dropout can be interpreted as a Bayesian neural network \cite{gal2016uncertainty}. 
Stochastic forward passes with dropout `turned-on' at test time then correspond to draws from the predictive distribution.
Here we extend on these ideas and propose the \textit{Variational Transformer}, which is based on MC Dropout applied to the original Transformer model. We perform extensive empirical evaluation with this model on the task of NMT.
Representing uncertainty in a translation model is the first step towards detecting OOD data. We next discuss how to \textit{use} this uncertainty effectively, and provide the main contribution of this work.

\section{Measures of Uncertainty for NMT}
\label{sect:measures}

Principally, we care about measuring the variance of a model's outputs around some given input point. In the context of a simple classifier model, the solution is often found by measuring the mutual information between the predicted discrete distribution and model parameters, evaluating the output's entropy, or simply computing the variance of model outputs \citep{gal2016uncertainty}. In the domain of language, however, there are many semantically equivalent alternatives to the same prediction, and it is a difficult matter to measure the disagreement between the predicted discrete sequences, which in turn complicates the estimation of variance in the output space.
Much worse, when attempting to naively use MI or entropy with long sequences or large sets of discrete random variables, we quickly discover that even approximate integration over the product space becomes prohibitive \citep{kirsch2019batchbald}.
%
% When computing a variance over outputs we are concerned with measuring an \emph{expected distance from the mean}; in the case of discrete sequences, the `mean' predicted sequence is difficult to compute (given a decoder's reliance on discrete sampling) and notions of a `distance' from this mean are ambiguous at best. 

In order to capture epistemic uncertainty in the task of NMT, we propose several measures of uncertainty appropriate for long sequences of discrete variables (Beam Score and Sequence Probability are measures similar to \citep{WangLWLS19}):
% In order to evaluate the effect different choices of confidence scores have on calibration we evaluate three alternatives:
\begin{enumerate}%[leftmargin=*]
    \item \textbf{Beam Score:} we assign a confidence to output \(y\) generated (using beam search) from input \(x\) using the score assigned to \(y\)'s beam \citep{wu2016google}, where 
    $\operatorname{length\_penalty}(y;\alpha) = \left(\frac{5 + |y|}{5+1}\right)^\alpha$.
        \begin{equation}\label{eqn:bs}
           \operatorname{BS} =  \frac{\log \left(p_{\omega^*}(y|x)\right) }{ \operatorname{length\_penalty}(y;0.6)}
        \end{equation}
        % \begin{equation}
        %     \operatorname{length\_penalty}(y;\alpha) = \left(\frac{5 + |y|}{5+1}\right)^\alpha
        % \end{equation}
    \item \textbf{Sequence Probability:} we assign a confidence to the deterministic model output \(y\) generated from input \(x\) by taking the log predictive probability under the weight distribution.
        \begin{align}\label{eqn:sp}
            \operatorname{SP} = \frac{\log \left( \mathbb{E}_{\omega\sim q_{\theta^*}(\omega)} \, p_\omega(y|x) \right) }{ \operatorname{length\_penalty}(y;0.6)}
        \end{align}
    \item \textbf{BLEU Variance:} 
    % \new{
        ideally, we would like to measure the variance of outputs $y$ as the uncertainty at an input $x$, i.e.:
        \begin{align}
            \operatorname{Var} = \mathbb{E}_{\omega\sim q_{\theta^*}(\omega)}\mathbb{E}_{y\sim p_\omega(y|x)} \left(y - \mu \right)^2 = \mathbb{E}_{\omega\sim q_{\theta^*}(\omega)}\mathbb{E}_{y, y'\sim p_\omega(y|x)} \frac{1}{2} \left(y - y' \right)^2,
        \end{align}
        where $\mu = \mathbb{E}_{y\sim p_\omega(y|x)} [y]$. If we treat sentences as points in some high dimensional space, $||y-y'||$ corresponds to a distance between these two points, which is a numerical value representing the difference between two sentences. Thus, any metric for measuring the difference between sentences will allow us to calculate the variance of output $y$. In our experiments we choose BLEU \citep{papineni2002bleu}. The BLEU score of a candidate text to the reference text is a number between 0 and 1, with the value closer to 1 indicating the two texts are more similar\footnote{A common practice in the NMT literature is to scale it up by $\times 100$, i.e. in the range of $[0,100]$.}.
        Now, we can estimate the variance at an input \(x\) by producing pairs of outputs from the model and measuring the squared complement of the BLEU between them, i.e. $||y-y'||^2 \coloneqq \left(1-\operatorname{BLEU}(y, y')\right)^2$, and we have:
            \begin{align}\label{eqn:bleuvar}
                \operatorname{BLEUVar} = \mathbb{E}_{\omega\sim q_{\theta^*}(\omega)}\mathbb{E}_{y,y'\sim p_\omega(y|x)} \left(1-\operatorname{BLEU}(y, y')\right)^2
            \end{align}
    % }
    
\end{enumerate}

For the \emph{beam score} we use the deterministic model found by gradient descent and simply take the probabilities from under its output probabilities. This will be our baseline. 
% For \emph{beam score} we use the predictive probabilities for the search by averaging multiple stochastic forward passes at each token evaluation. 
For \emph{sequence probability} and \emph{BLEU variance} we use MC Dropout \citep{gal2016uncertainty} and take a number of samples (\(N\)) to estimate the expectations\footnote{The approximations below should have constant scaling factors, but these don't impact our evaluation metric (performance-retention curves) so we leave them out for simplicity.}:
\begin{equation}
    \operatorname{SP} \approx \frac{\log  \left( \sum^N_{i=1} p_{\omega_i}(y|x) \right) }{ \operatorname{length\_penalty}(y;0.6)}
\end{equation}
For the \(\operatorname{BLEUVar}\) approximation, we opt for decoding outputs using beam search applied to different model samples (realised by randomising the dropout masks) and measuring the complement BLEU between pairs of these examples.
\begin{align}
    \operatorname{BLEUVar} \approx \sum^N_{i=1} \sum^N_{j \neq i} \left(1-\operatorname{BLEU}(\operatorname{dec}_{\omega_i}(x), \operatorname{dec}_{\omega_j}(x))\right)^2.
\end{align}
Additionally, out of the $N$ sample sequences generated by BLEUVar, we need to choose one or generate a new sequence as the result for a specific input. In regression, the mean of the samples is normally chosen as the result, which is an approximation for the \emph{predictive mean}. In our case, the `mean' of $N$ sentences is hard to derive or even not properly defined. Therefore, we use the sampled sequence that is the closest to the ideal `mean' as an approximation. Since we can measure the disagreement between any two sentences using BLEU, the sequence that is the closest to the `mean' of the $N$ samples must have the smallest disagreement with rest of the $N-1$ samples. Hence, the final output sequence of the method BLEU Variance is:
\begin{align}
    \tilde{\mu} = 
    \argmin_{{y}_i} \Big( \sum_{\forall j \neq i}^N \big(1-\operatorname{BLEU}(y_i, y'_j)\big) + \sum_{\forall j \neq i}^N \big(1-\operatorname{BLEU}(y'_j, y_i)\big)\Big).
\end{align}
% \tim{Need review $\rightarrow$} 
One remark for the above three methods is that the two methods BS and SP have the same resulting translation $y$ given $x$, but with different values as its uncertainty estimates. For the method BLEUVar, it uses $\tilde{\mu}$ for the resulting translation, and a value in a different range as the uncertainty estimate. This is reflected in the plots from the experiments section. For example, in Figure \ref{fig:envi-350k-mc10-50} (b), BS and SP converged into the same value, while BLEUVar converged to a different value.

\subsection{Evaluating Uncertainty in Sequence Models}

A number of uncertainty evaluation metrics have been proposed for standard classifier networks such as the popular ECE and MSE metrics \citep{guo2017calibration}; however, for sequence modelling these classification-specific metrics are not applicable. Instead, we opt for the \emph{performance versus retention} curve method for evaluating our uncertainty measures following \citet{oatml2019bdlb}.

The performance-retention curve indicates how well an uncertainty measure would perform if the \(k\)\% \emph{least} certain outputs were deleted from the test dataset. The \(x\)-axis ranges along the fraction of data retained, while the \(y\)-axis measures some performance metric of the model on the retained data.
A performance-retention curve of a well-calibrated 
uncertainty measure will see a clear and sustained improvement in performance as low-confidence predictions are excluded from the test set; while a poorly-calibrated model will yield a curve that either lacks a trend or tends to lay beneath the well-calibrated metric's curve.

\section{Experiments}
\label{sect:exps}

% \acl{
%     Move the similar plots to appendix
% }

Experiments consist of evaluations on both in-distribution (see Appendix \ref{sect:in-dist}) and out-of-distribution test sets. 
The implementation of the Transformer architecture we use is taken from the Tensor2Tensor \citep{tensor2tensor} repository.

As discussed in the previous sections we use performance-retention curves to evaluate the different uncertainty estimates of our models. We also use scatter plots of the uncertainty versus pairwise BLEU of the predictions to offer another gauge of model uncertainty (see Appendix \ref{sect:train-size})
To visualise the quality of uncertainty estimates for Transformers, we can order the generated sequence based on their uncertainty estimates from the most confident to the most uncertain, and plot BLEU scores as a function of the fraction of data retained starting from the least uncertain output. For a good uncertainty estimate, we are expecting to see the BLEU scores decrease when the fraction of retained data increases.

The following datasets were used in our experiments:
%  \textbf{(1) WMT EN $\leftrightarrow$ DE}: The training set for translation tasks between English (EN) and German (DE) composed of \emph{news-commentary-v13} with $284k$ sentences pairs, \emph{wmt13-commoncrawl} with $2.4m$ sentences pairs and \emph{europarl-v7} with $1.9m$ sentences pairs, in total $4.6m$ sentences pairs. The test set was the \emph{newstest2014} with size $3k$ from \emph{WMT 2014}.
%  \textbf{(2) WMT NL $\rightarrow$ EN}: The test set for Dutch (NL) to English (EN) translation was a subset (size $3k$ sentences pairs) of \emph{news-commentary-v14}.
%
\begin{itemize}%[leftmargin=*]
    \item \textbf{WMT EN $\leftrightarrow$ DE}: The training set for translation tasks between English (EN) and German (DE) composed of \textbf{news-commentary-v13} with $284k$ sentences pairs, \textbf{wmt13-commoncrawl} with $2.4m$ sentences pairs and \textbf{europarl-v7} with $1.9m$ sentences pairs, in total $4.6m$ sentences pairs. The test set was the \textbf{newstest2014} with size $3k$ from \textbf{WMT 2014}. 
    
    \item \textbf{WMT NL $\rightarrow$ EN}: The test set for Dutch (NL) to English (EN) translation was a subset (size $3k$ sentences pairs) of \textbf{news-commentary-v14}.
\end{itemize}

In the experiments below, we denote the results from methods \emph{beam score}, \emph{sequence probability}, and \emph{BLEU Variance} as \emph{BS}, \emph{SP} and \emph{BLEUVar} respectively. If a suffix is added such as \emph{BLEUVar-10}, then the suffix \emph{10} indicates the number of samples taken during MC dropout.

\subsection{Out-of-Distribution Experiments}

The in-distribution tests showed in Appendix \ref{sect:in-dist} illustrate that when the training set is large enough, the three uncertainty estimates have similar performance if the test set is in the same domain as the training set. However, when the training set only has limited data (e.g. $50k$ compares to $4.6m$), BLEUVar outperforms BS and SP even though the test set is in the same domain as the training set. One explanation is that with the limited amount of training data, the model was not able to learn a distribution that capture the data from the test set. Therefore, to some extend, the test data are slightly out-of-distribution.

In this section, we will have a look of experiments under two different obviously out-of-distribution settings (which in our application is equivalent to a \textit{domain shift}) for evaluating our uncertainty estimates. 

The first experiment exploits the fact that the wordings and the sentence patterns might vary across different content domains in the same language. For example, sentences from a legal document and sentences from social media are different in terms of formality, even though they can both be in the same language. If we train a model using data from one domain and test it with data from another domain, then such input would be out-of-distribution for the model.

The second experiment explores an extreme case of out-of-distribution setting. Given a trained model for translation task from language $A$ to language $B$, if we test it on input from language $C$ s.t. $C\neq A$, it would be an out-of-distribution input.

% The test sets were in different domain than the training sets.

\subsubsection{Uncertainty Caused by Different Content Domains}

% \tim{modify the dataset spec info before}

\begin{figure*}[ht!]
% \vspace{-1em}
    \centering
    \begin{subfigure}[b]{0.4\textwidth}
        \centering
        \includegraphics[width=\textwidth]{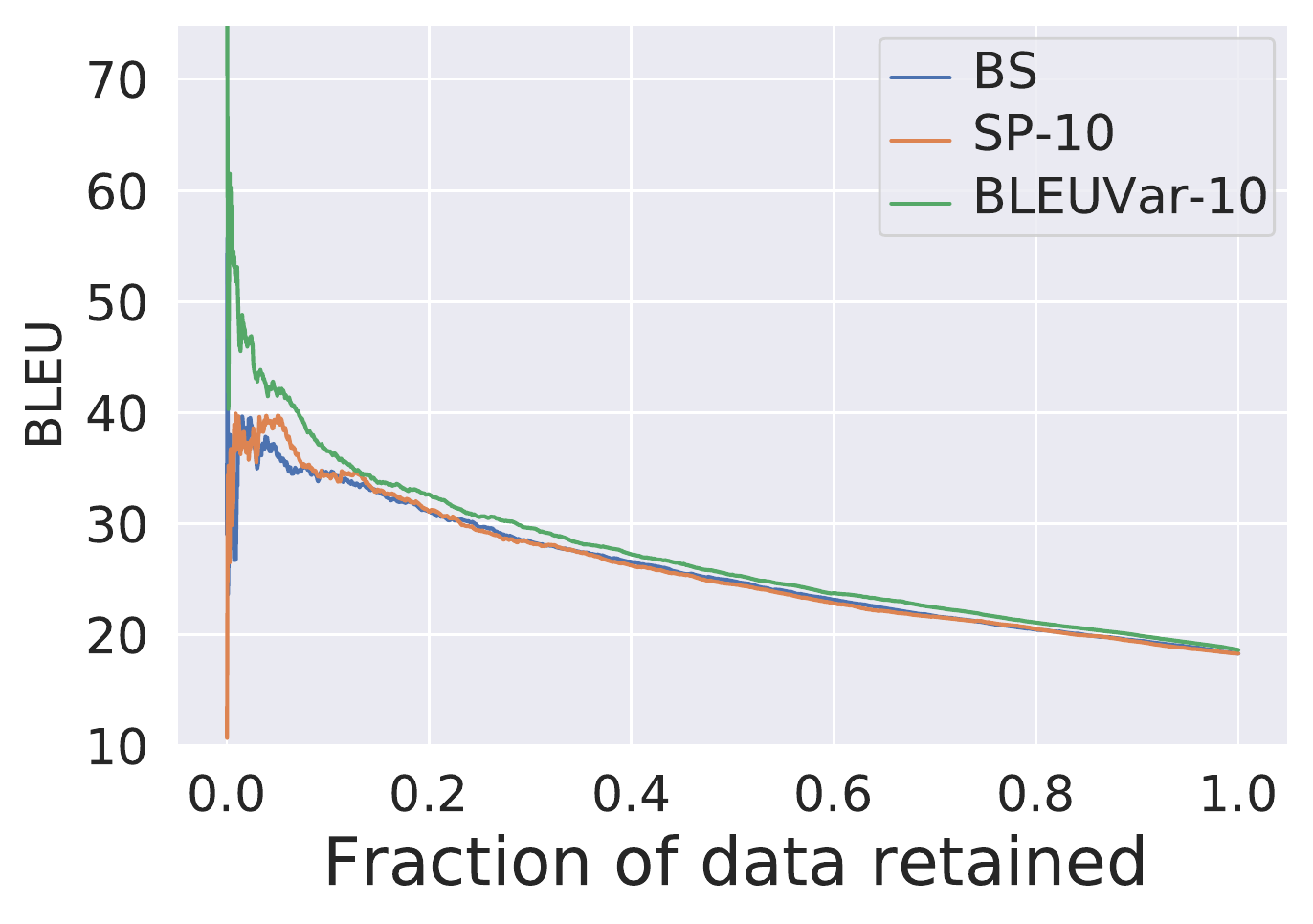}
        \caption{In-domain test set \emph{newstest2014} (size $3k$)}
    \end{subfigure}
    \begin{subfigure}[b]{0.4\textwidth}
        \centering
        \includegraphics[width=\textwidth]{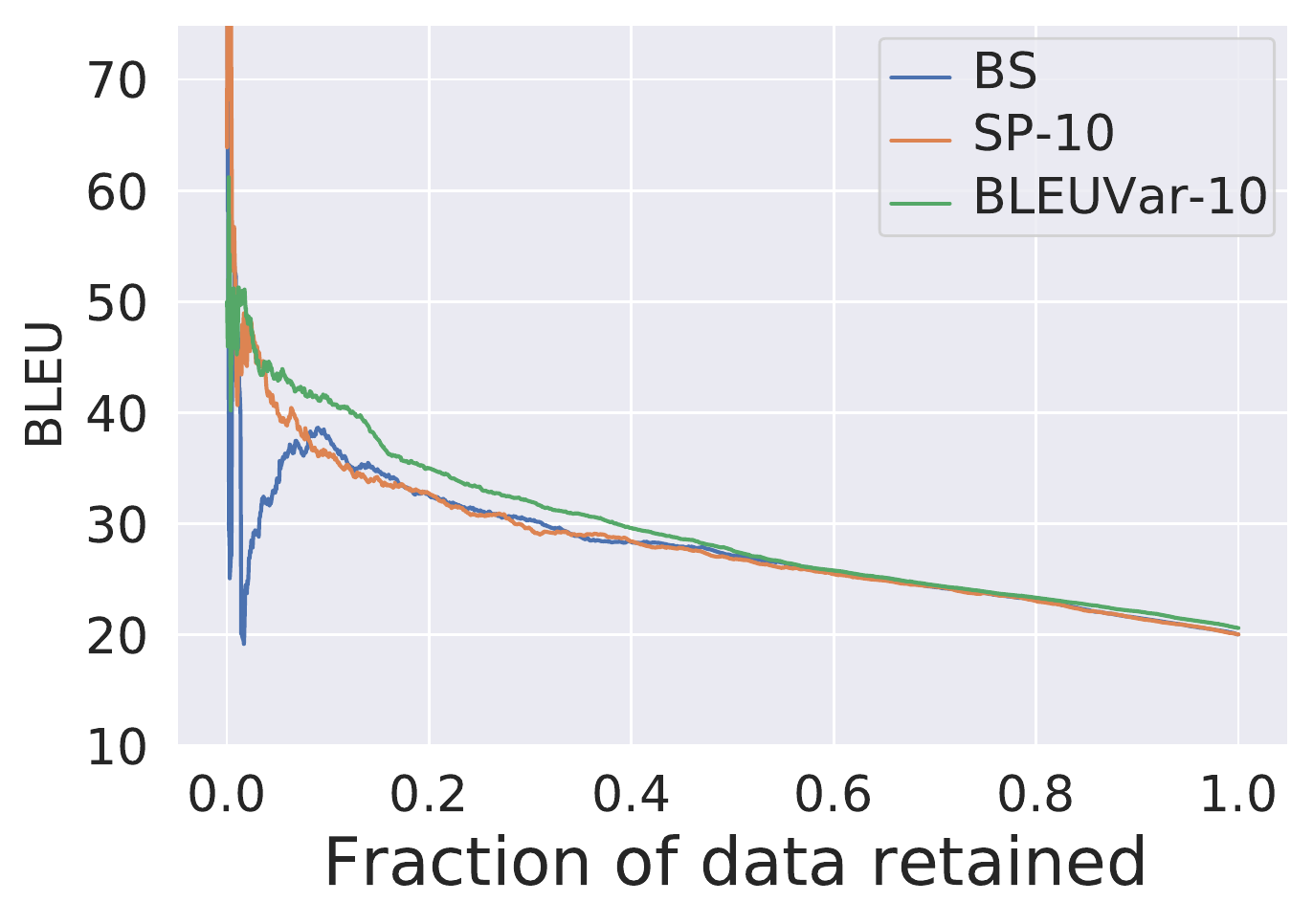}
        \caption{Out-of-domain test set \emph{Europarl} (size $3k$)}
    \end{subfigure}
    \caption{Uncertainty measure comparisons using the same-domain test set \emph{newstest2014} (left) and out-of-training-domain test set \emph{Europarl} (right). The Transformer model was trained for DE to EN tasks with the \emph{news-commentary-v13} EN-DE training set (size $284k$) using $350k$ steps.}
    \label{fig:de2en-nc-nt-europarl}
    % \vspace{-1em}
\end{figure*}

To evaluate our uncertainty measures on out-of-content-domain test set, we trained a model using only the \emph{news-commentary-v13} data for German to English (DE-EN) task. This training set has $284k$ samples in the domain of news commentary. During test time, we used both the in-domain test data (\emph{newstest2014}) and the out-of-domain test data (a subset of \emph{Europarl}). \emph{newstest2014} is in the same domain as the training set. \emph{Europarl} contains samples extracted from the proceedings of the European Parliament, which has a different domain than news commentary.

We expect a good uncertainty measure to perform well in both in-domain and out-of-domain test set. As shown in Figure \ref{fig:de2en-nc-nt-europarl}, BLEUVar outperforms the other two measures by a large margin for both test sets. In particular, the unstable performance of BS on the out-of-domain test data indicating such uncertainty measure might be not be reliable on out-of-distribution data, i.e.\ data it never saw before.

% \subsubsection{EN to DE trained without news data}
% In this experiment, we only used the data set \textit{wmt13-commoncrawl} as the training set, which does not contain news data. And our test set was \textbf{newstest2014}, which only contains news data, therefore OOD.

% \begin{figure}[ht!]
%     \centering
%     \includegraphics[width=0.48\textwidth]{sb-ende-350k-CC-mc10.png}
%     \caption{Uncertainty estimator comparisons for the model trained only using \textit{wmt13-commoncrawl}. The number of training steps was 350k, and the task is EN to DE.}
%     \label{fig:ende-350k-CC-mc10}
% \end{figure}

% Take away:

% \begin{itemize}
%     \item The method \emph{BLEU variance} outperforms \emph{beam score} and \emph{sequence probability}.
% \end{itemize}

% \subsubsection{Detecting Out of Language Domain Sentences in NMT}
\subsubsection{Uncertainty Caused by Different Language Domains}
\label{sect:ood}
% Training data was the full EN-DE training set (size $4.6m$). The task was German to English translation.

% Test set: $3k$ of DE to EN test data from \textit{newstest2014} combined with $3k$ of NL to EN test data from \textit{news-commentary-v14} resulted in $6k$ test data in total. The experiments were performed using the DE+NL test set and the DE to EN model.

\begin{figure*}[ht!]
% \vspace{-1.3em}
    \centering
    \begin{subfigure}[t]{0.33\textwidth}
        \centering
        \includegraphics[width=\textwidth]{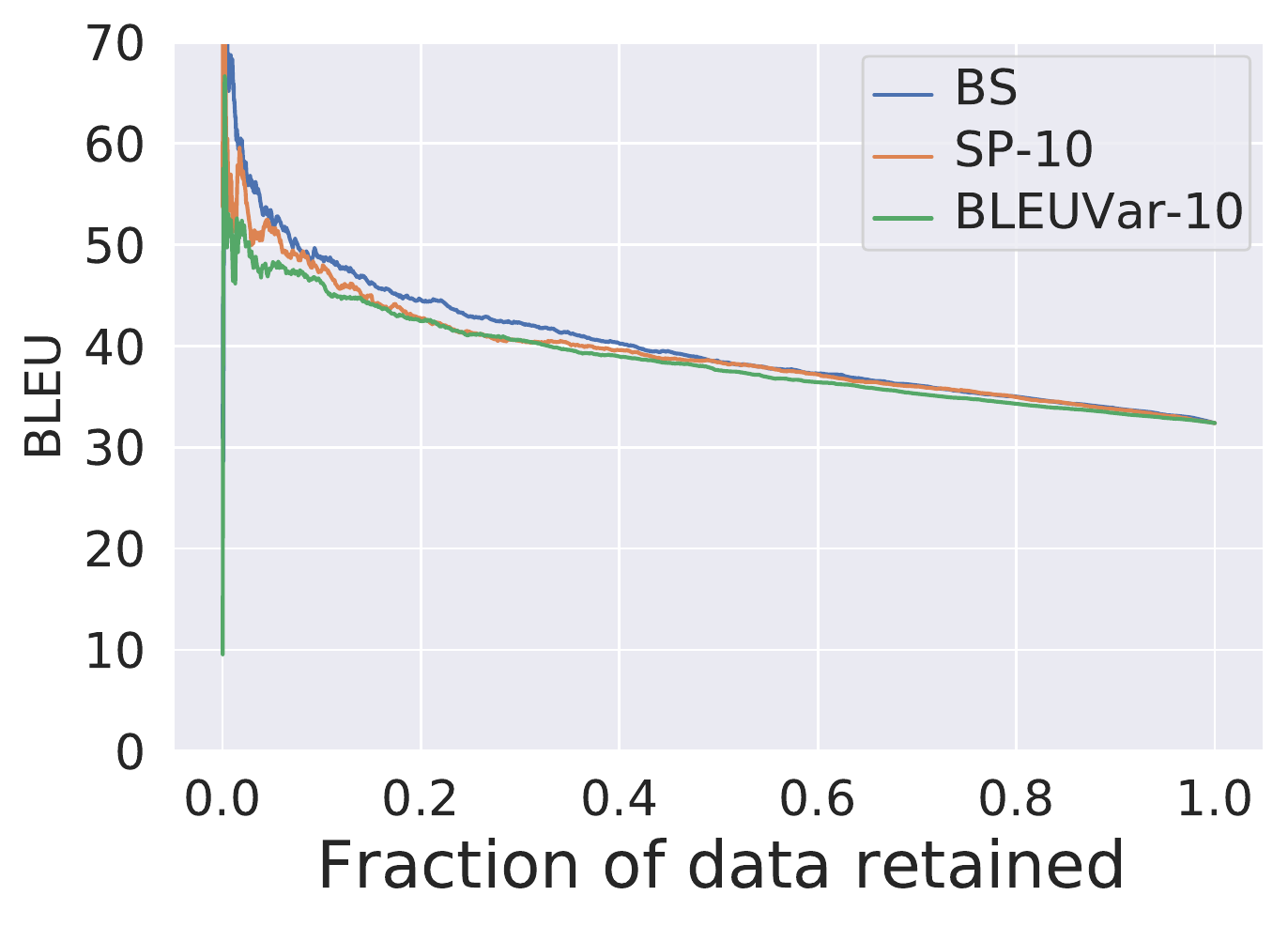}
        \caption{DE to EN (size $3k$)}
    \end{subfigure}%\hspace{0.03\textwidth}
    \begin{subfigure}[t]{0.33\textwidth}
        \centering
        \includegraphics[width=\textwidth]{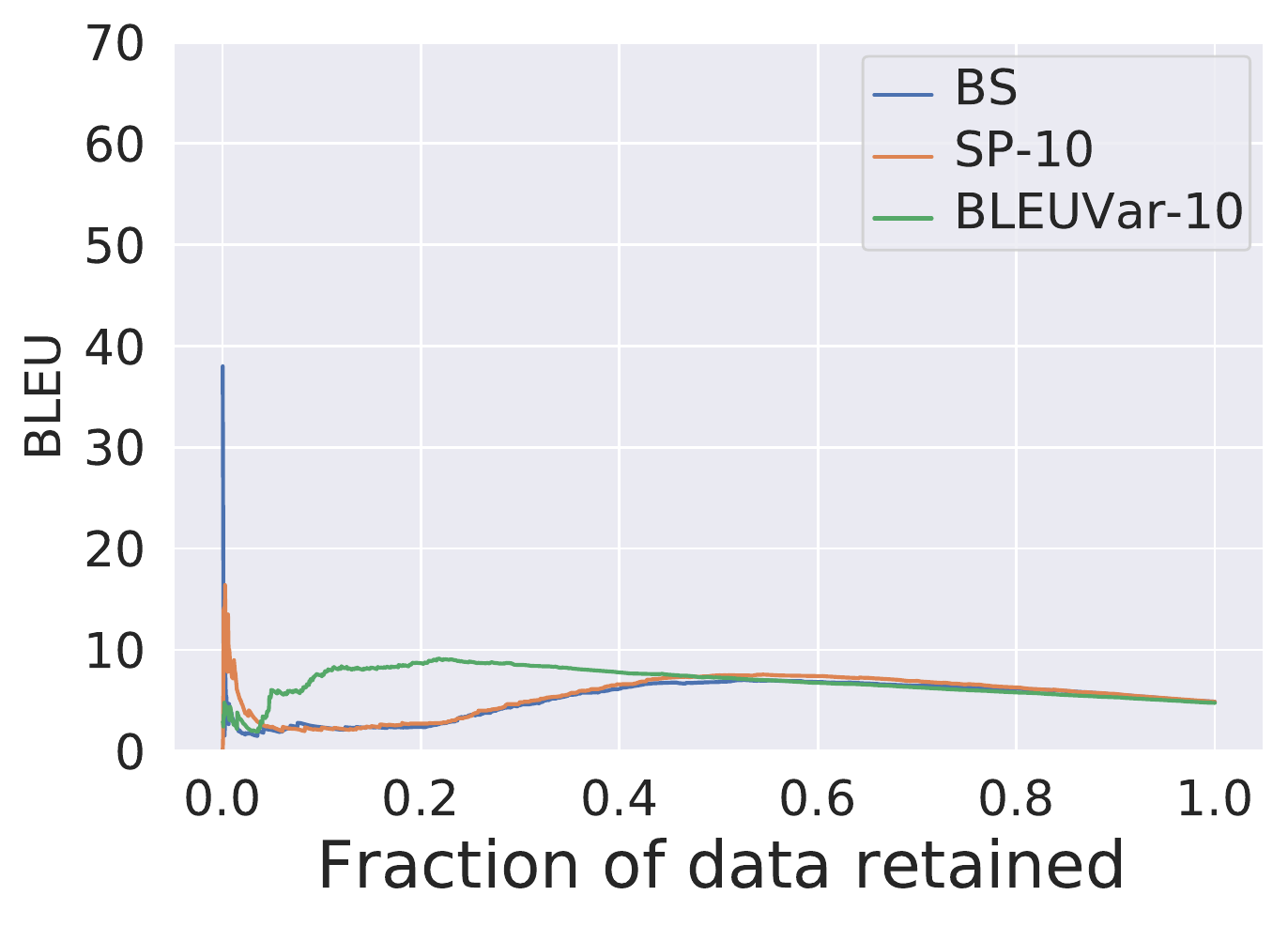}
        \caption{NL to EN (size $3k$)}
    \end{subfigure}%\hspace{0.03\textwidth}
    \begin{subfigure}[t]{0.33\textwidth}
        \centering
        \includegraphics[width=\textwidth]{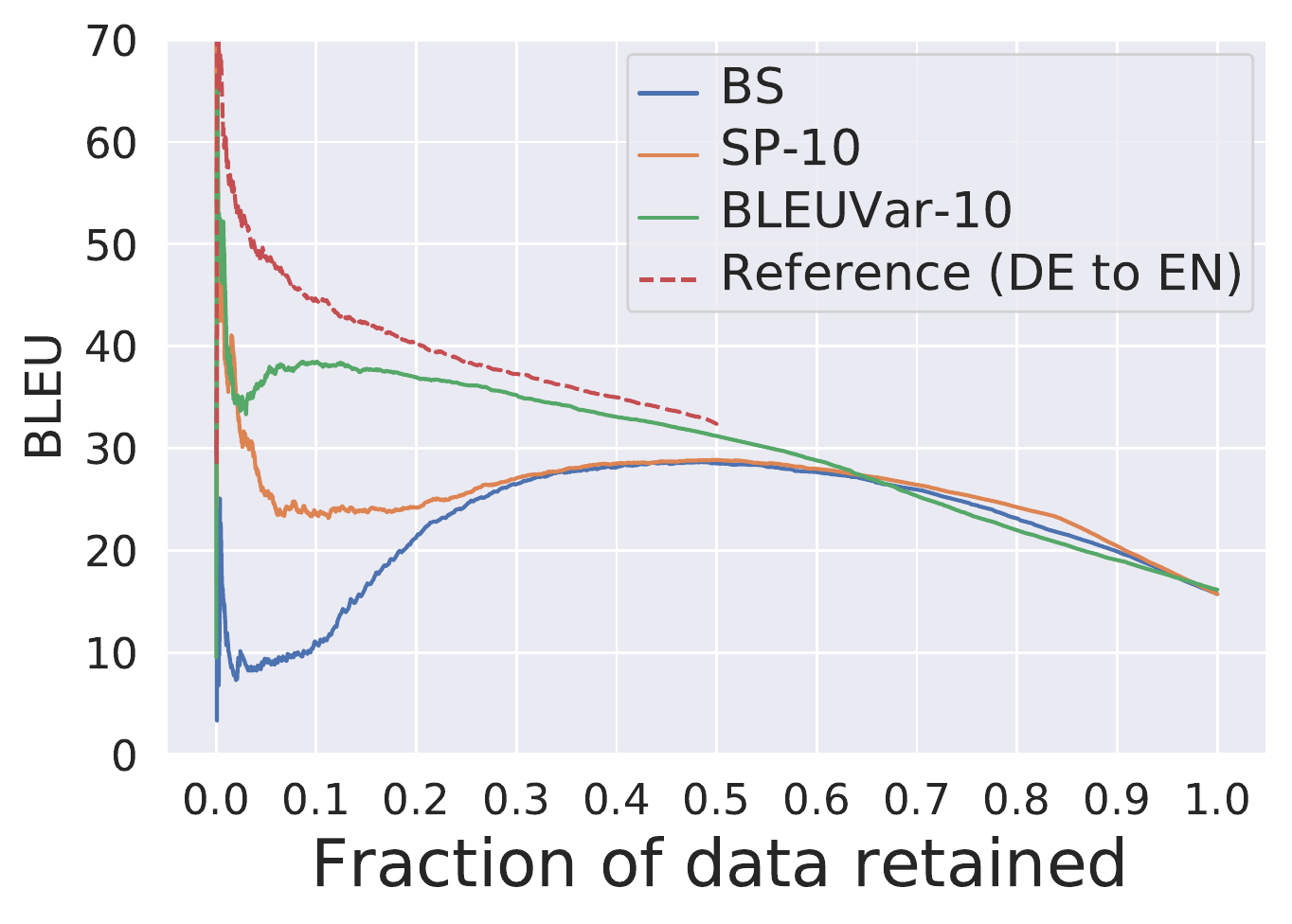}
        \caption{DE+NL to EN (size $6k$)}
    \end{subfigure}
    \caption{Uncertainty measure comparisons using the in-distribution DE-EN test set (a), out-of-distribution NL-EN test set (b) and the combined DE+NL to EN test set (c). The \emph{Reference} line in (c) corresponds to the BS plot from (a), which only has $3k$ test data. Therefore it only reaches the fraction 0.5 in this graph. The model was trained for DE to EN task with the full EN-DE training set (size $4.6m$) using $350k$ steps.}
    \label{fig:de-en-nl-en}
    % \vspace{-1em}
\end{figure*}

We trained a Transformer model on WMT13 and Europarl DE (German) to EN (English) sentence pairs (obtaining BLEU 33 on the WMT14 test set). We then evaluated the model on out-of-training-distribution input sentences in NL (Dutch), which shares a large overlapping vocabulary with German (hence input sentences look plausible to non-native speakers). One would hope that such data falling outside of the training distribution would produce model predictions with high uncertainty.

%  To evaluate Figure \ref{fig:de+nl-en}: If the uncertainty estimate has high quality, the left-hand side of the plot (before fraction 0.5) generated using the combined DE+NL test set will be very close to the plot generated only with the DE test set (i.e. the plot in Figure \ref{fig:de-en-nl-en} (a)). This is because a high-quality uncertainty estimate will order the results from the combined DE+NL test set based on their uncertainty such that the in-distribution data (i.e. DE), which are assigned with lower uncertainties, will be placed on the left half of the plot.

The BLEU scores differences between Figure \ref{fig:de-en-nl-en}(a) and (b) shows that feeding Dutch (NL) sentences into a German to English model does not result in meaningful translations in general. Nevertheless, BLEUVar still provide a better uncertainty estimate with Dutch input (see Figure \ref{fig:de-en-nl-en}(b)). 

In addition, Figure \ref{fig:de-en-nl-en}(c) shows the performance-retention curves for the combined DE+NL test set, in which BLEUVar outperforms BS and SP by a large margin. The fact that BLEUVar is close to the DE-EN reference curve indicates most of the DE input has been assigned with high confidence correctly. As the left half of the BLEUVar curve, which corresponds to the most certain half of the test data, nearly resembles the result from Figure \ref{fig:de-en-nl-en}(a).     

\begin{figure*}[ht!]
% \vspace{-1em}
    \centering
    \begin{subfigure}[b]{0.4\textwidth}
        \centering
        \includegraphics[width=\textwidth]{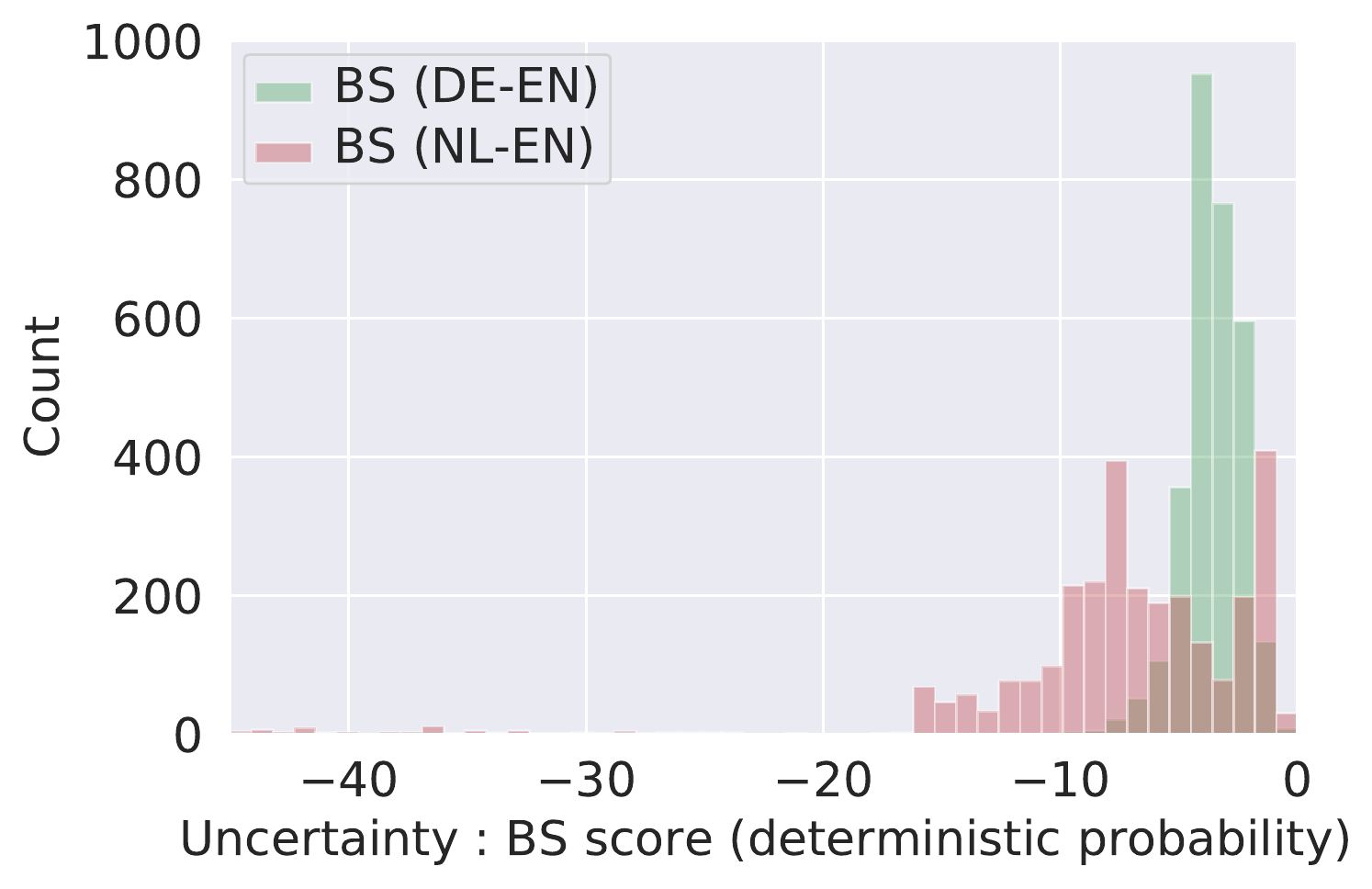}
        \caption{BS histogram on a mixed test set (Left corresponds to higher uncertainty)}
    \end{subfigure}\hspace{0.03\textwidth}
    \begin{subfigure}[b]{0.4\textwidth}
        \centering
        \includegraphics[width=\textwidth]{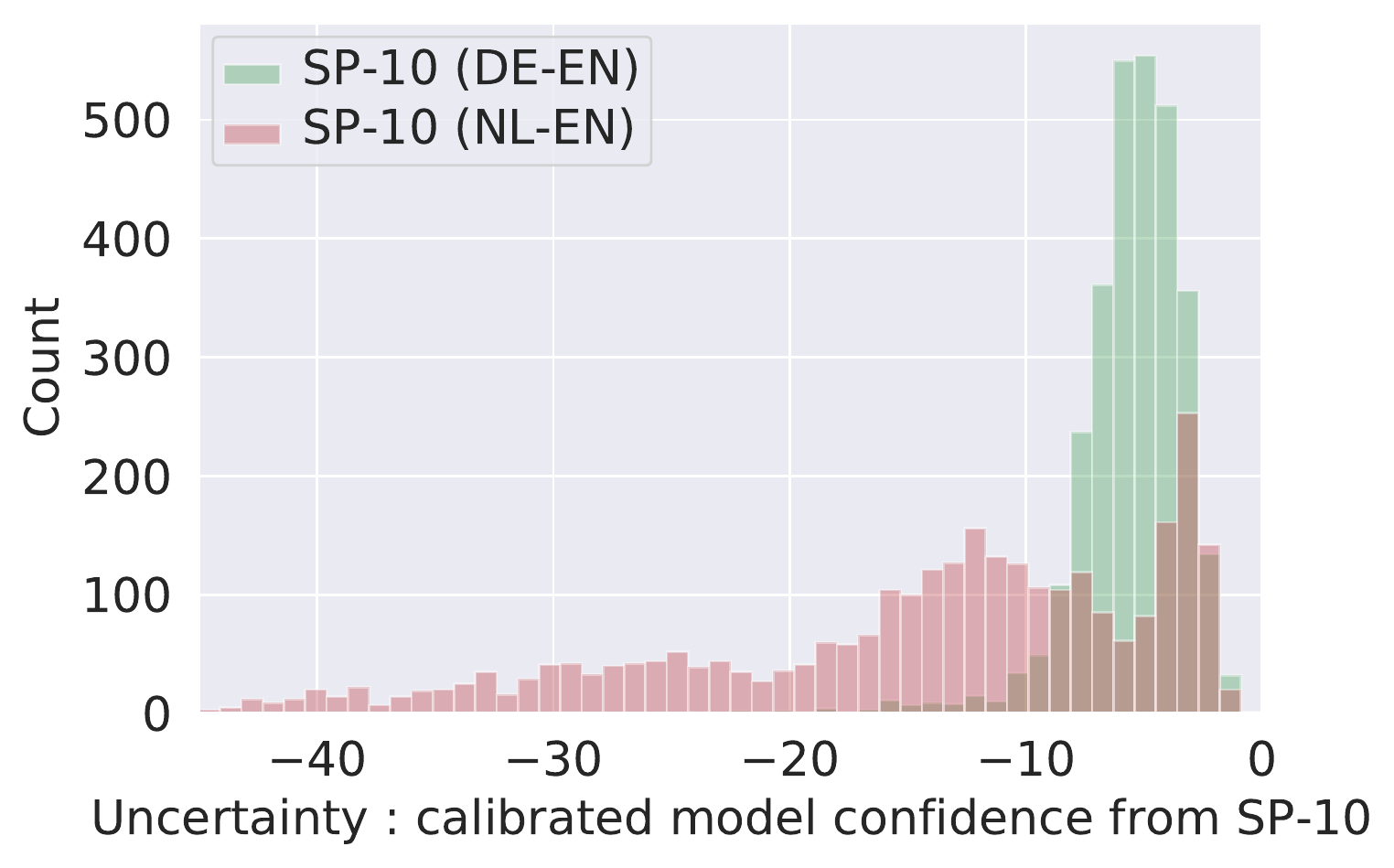}
        \caption{SP-10 histogram on a mixed test set (Left corresponds to higher uncertainty)}
    \end{subfigure}\hspace{0.03\textwidth}
    \begin{subfigure}[b]{0.4\textwidth}
        \centering
        \includegraphics[width=\textwidth]{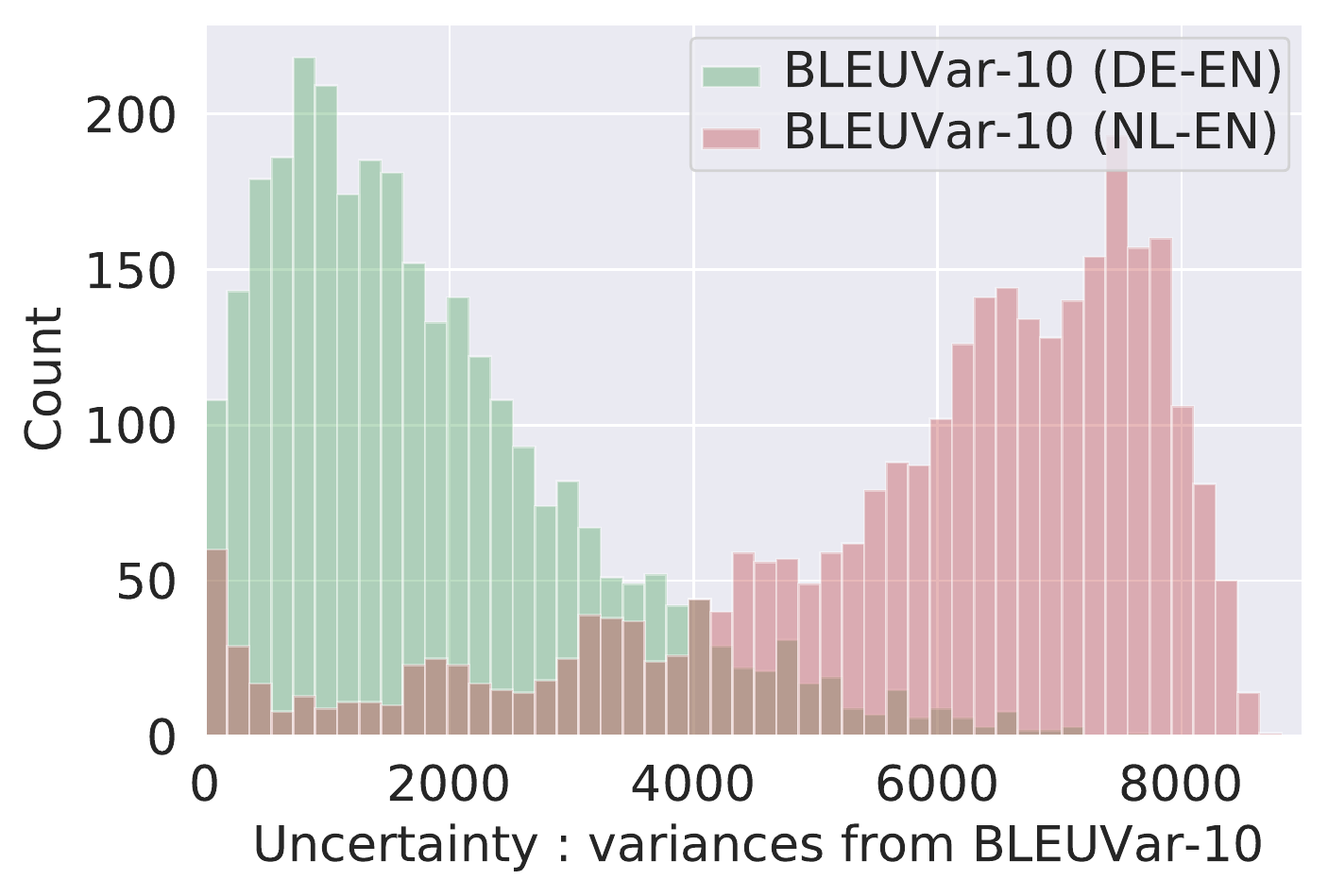}
        \caption{BLEUVar-10 histogram\protect\footnotemark \ on a mixed test set (Right corresponds to higher uncertainty)}
    \end{subfigure}
    \caption{
    Histograms show uncertainty value for DE-EN (green) and NL-EN (red). Note how BLEUVar-10 is able to clearly separate the in-distribution (green) from out-of-distribution (red). The model was trained for DE to EN task with the full EN-DE training set (size $4.6m$) using $350k$ steps.}
    \label{fig:hists}
    % \vspace{-1em}
\end{figure*}

A more interesting result is shown in Figure \ref{fig:hists}. Given the combined DE+NL test set, BLEUVar is able to nicely separate the test set into to two cluster (DE and NL) using only the uncertainty estimates without evaluating on the target translation. 

% \begin{figure*}[hb!]
%     \centering
%     \begin{subfigure}[b]{0.48\textwidth}
%         \centering
%         \includegraphics[width=\textwidth]{sb-de-en.png}
%         \caption{DE to EN (size $3k$)}
%     \end{subfigure}
%     \begin{subfigure}[b]{0.48\textwidth}
%         \centering
%         \includegraphics[width=\textwidth]{sb-nl-en.png}
%         \caption{NL to EN (size $3k$)}
%     \end{subfigure}
%     \caption{Uncertainty measure comparisons using the in-distribution DE-EN test set (left) and out-of-distribution NL-EN test set (right). The Transformer model was trained for DE to EN tasks with the full EN-DE training set (size $4.6m$) using $350k$ steps.}
%     \label{fig:de-en-nl-en}
% \end{figure*}

% \tim{add more description here}

% \begin{figure}[ht!]
%     \centering
%     \includegraphics[width=0.45\textwidth]{sb-de+nl-en.png}
%     \caption{Comparison for different uncertainty measures on a combined DE+NL test set ($6k$ size). The \emph{Reference} line corresponds to the BS plot from the previous Figure \ref{fig:de-en-nl-en} (a), which only has $3k$ test data. Therefore it only reaches the fraction 0.5 in this graph. The model was trained for DE to EN task with the full EN-DE training set (size $4.6m$) using $350k$ steps.}
%     \label{fig:de+nl-en}
% \end{figure}

\footnotetext{With the common practice of BLEU (i.e. $\times 100$), the BLEUVar value results in $\times 100^2$ in the plots.}

% \begin{figure}[ht!]
%     \centering
%     \includegraphics[width=0.3\textwidth]{sb-de+nl-en-individual-hex.png}
%     \caption{The density of individual BLEU scores for sentences in the DE+NL to EN test set. The sentences are ordered by their uncertainty from low (left) to high (right) using \textit{BLEUVar-10}. The model was trained for DE to EN tasks with $4.6m$ training data using $350k$ steps.}
%     \label{fig:de+nl-en-individual}
% \end{figure}

% Take away:

% \begin{itemize}
%     \item The method BLEUVar outperforms SP, which outperforms BS.
%     \item The method BLEUVar can separate DE+NL into DE and NL with relatively high precision.
% \end{itemize}

% \new{

Note that we do not try to solve the problem of language detection here. Instead, we use it as an extreme example to show that our uncertainty metric is able to assign very high uncertainty scores towards out-of-distribution test data while being more confident towards in-distribution data. But our methodology was not trained specifically nor designed for language detection. However, being able to separate the two shows that our methods do reflect the uncertainty of the model very well, it also shows that Bayesian deep NLP models can be very powerful. 

\subsection{Analysis of Sentence Length versus Uncertainty}

Use sentences with different lengths and uncertainties from section \S\ref{sect:ood} as examples, we can see from the Table \ref{tab:lengths} that long sentences do not necessarily have consistently low/high uncertainty (BLEUVar). For the in-distribution test data, the model is very certain for sentences of length around 21-50, this is because the training sentences length distribution is centred around these sentence lengths (with 54\% of the training data in this interval). For the OOD test data, the model is very uncertain across all sentence lengths, and as the sentences get longer the model becomes more uncertain. Note that average uncertainty for the shortest OOD sentences is $\sim$4700, which is much greater than the average uncertainty for the longest in-distribution sentences, $\sim$2700.

\begin{table}[ht!]
% \vspace{-1em}
% \small
\centering
\begin{tabular}{l||c|c|c|c|c|c}
% \hline
Lengths & 1-10  & 11-20 & 21-30 & 31-40 & 41-50 & 51+ \Tstrut\Bstrut\\ \hline
DE-EN (In-dist)  & 2579.93 & 1867.11 & 1613.95 & 1507.85 & 1502.78 & 2794.11 \Tstrut\Bstrut\\
NL-EN (OOD)  & 4772.16 & 5388.25 & 5579.82 & 6039.02 & 6705.95 & 7042.69 \Tstrut\Bstrut\\
\end{tabular}

\caption{\Tstrut Average BLEUVar for output sentences of various lengths from Figure \ref{fig:de-en-nl-en}.}
\label{tab:lengths}
% \vspace{-1.5em}
\end{table}

An important point is that it is not necessary that \textit{all} the short sentences must have low uncertainty, even for sentences in-distribution. I.e., our uncertainty metric is not simply a measure of sentence lengths (or even correlated with it).

We further extended the experiment design and added a naive baseline which simply looks at the sentence length to reject sentences (ordering test data using $\frac{sentence\_length }{ longest\_sentence\_length}$, i.e. from short sentences to long sentences). Calculating BLEU scores at different retention rates (as in Figure \ref{fig:de-en-nl-en}(a), see Table \ref{tab:ret-len}), we see that the curve is much lower than BLEUVar. In fact, the performance of sentence length referral is worse than another naive baseline: randomly referring sentences without looking at them at all.

\begin{table}[ht!]
% \small
% \vspace{-0.5em}
\centering
\begin{tabular}{c|c|c|c}
% \hline
Retention & Sentence length & Random referral & BLEUVar \Tstrut\Bstrut\\ \hline
0.2 &       29.62 &           32.06 &           42.87 \Tstrut\Bstrut\\
0.3 &       31.48 &           31.93 &           41.28 \Tstrut\Bstrut\\
0.4 &       32.38 &           32.26 &           39.39 \Tstrut\Bstrut\\
\end{tabular}

\caption{\Tstrut BLEU scores at different retention rates under three ordering (sentence length here is system output sentence length; source sentence length behaves the same).}
\label{tab:ret-len}
% \vspace{-2em}
\end{table}

\subsection{Introspection into the Model Uncertainty from the Linguistic Perspective}

Based on experiments in section \S\ref{sect:ood}, below are some example sentence translations sampled from the model (and with which we estimate the model uncertainty). As a reminder, the uncertainty here (BLEUVar) is the level of disagreement between sampled sentences, as determined by pairwise-BLEU (pairwise between each model output and the other model outputs).
For a very confident translation from an in-distribution input (i.e. German), refer to Appendix \ref{sect:samples:in-dist-de-1} Table \ref{tab:in-dist-de-1}. We can see that all translations sampled from the model are consistent with each other, and the model has no uncertainty at all. 
For other less certain in-distribution translations as showed in in Appendix \ref{sect:samples:in-dist-de-2} Table \ref{tab:in-dist-de-2}, we can see that each translation sampled from the model is inconsistent with the others in subtle ways, leading to a larger variability in pair-wise BLEU scores. The model has high uncertainty, but still lower than that of the average OOD sentence. In contrast, Table \ref{tab:ood-nl-short} shows 3 truncated samples from OOD input (i.e. Dutch), the complete table with 5 full samples can be found in Appendix \ref{sect:samples:ood-nl}. Here each translation sampled from the model is wildly inconsistent with the others, with some translations reminiscent of nonsense translations often encountered with neural systems when these are run on inputs they never saw before. We can identify these bad translations by the large variability in pair-wise BLEU scores. The model has much higher uncertainty than that of the average in-distribution sentences.

% \begin{table}[ht!]
%   \begin{center}
\begin{small}
    \begin{longtable}{l|l|p{6cm}}
      \toprule % <-- Toprule here
      \multicolumn{3}{l}{\textbf{Source sentence} (NL) : } \\[0.3em]
      \multicolumn{3}{p{13cm}}{De debiteurenlanden zouden hun concurrentiekracht terugkrijgen; hun schulden zouden in reële termen afnemen; de dreiging van staatsbankroeten zou - met de ECB onder hun controle - verdwijnen, en hun leenkosten zouden dalen naar een niveau dat vergelijkbaar is met dat van het Verenigd Koninkrijk.}\\
      \midrule % <-- Midrule here
      \multicolumn{3}{l}{\textbf{Reference translation} (EN) : {\small(only used to compute “BLEU to reference”)
}} \\[0.3em]
      \multicolumn{3}{p{13cm}}{Debtor countries would regain their competitiveness; their debt would diminish in real terms; and, with the ECB under their control, the threat of default would disappear and their borrowing costs would fall to levels comparable to that in the United Kingdom.}\\
      \midrule % <-- Midrule here
      \multicolumn{3}{l}{\textbf{Model predictive-mean translation} (EN) : {\small(averaging over predictive probabilities during decoding)
}} \\[0.3em] 
    \multicolumn{3}{p{13cm}}{The debitenlands were to compete with the rivalrivalrivalrivalrivalrivalrivalrivals of terugkrijgen; they were in debt in the countries of \seqsplit{afafafafafafafafafafafafafafafafafafafafafafafafafaf}}\\
      \midrule % <-- Midrule here
      \multicolumn{2}{l|}{\textbf{Translation “BLEU to reference”} : } & 1.9\\
      \midrule % <-- Midrule here
      \multicolumn{2}{l|}{\textbf{Translation uncertainty} : } & 8617\\
      \midrule % <-- Midrule here
      
      \multicolumn{3}{l}{\textbf{Translations sampled from the model}: {\small(3 shorten samples from predictive probabilities during decoding)}} \\
      \midrule % <-- Midrule here
      \textbf{1} & \multicolumn{2}{p{13cm}}{The debitenlands were the ones to compete in their rivalrivalrivalrivalrivalrivalrivals of them; they were debt-denominated in their afafafafafafafafen; the tripthirthirthirwent of state bankrbankrbankrbankru - with the ECB in its control the run - the run-off - the run - the run-run run run run run of the ECB.} \\
      \textbf{2} & \multicolumn{2}{p{13cm}}{In the debdebdebdebdebits were competitive in terms of law; those debt owed in debt in debt; the three of state \seqsplit{bankrbankrbankr} -  with the ECB, in its, in its, in its, in its control - business - business - the ECB, in its, in the control - the dispute, the - business - the dispute, the dispute, } \\
      \textbf{3} & \multicolumn{2}{p{13cm}}{At the time of its independence, it was a rivalrivalrivalrivalrivalrivaleach one; the debts of the poor; the three-three of the bankrbankrbankrall - with the ECB in its control of the ones - the ones in question - the parties in question, the countries in the future; the three of the bankrbankrbankrbankrbankr } \\

      \bottomrule % <-- Bottomrule here
    % \end{tabular}
    \caption{\Tstrut Out-of-distribution NL source sentence from the experiment in Figure \ref{fig:de-en-nl-en}(c). }
    \label{tab:ood-nl-short}
    \end{longtable}
%   \end{center}
\end{small}

% }

% \section{Conclusion and Future Directions}
% \label{sect:concs}

% In this work we propose and evaluate the different measures of uncertainty in NLP. We find empirically that deterministic Transformer baselines tend to place remarkably high confidence on their outputs for a previously unseen language (while performing very poorly in terms of BLEU). 
% We find that performing multiple stochastic forward passes through Transformers with dropout enabled at test time and evaluating the averaged probabilities -- that is, using Eq.(\ref{eqn:sp}) -- results in markedly improved uncertainty estimation. And even more drastic improvement is found by using our proposed \(\operatorname{BLEUVar}\) uncertainty estimation technique (Eq.(\ref{eqn:bleuvar})); \(\operatorname{BLEUVar}\) produces a substantial improvement in the separation between in-distribution and out-of-distribution sentences (Figure \ref{fig:hists}(c)). However, one drawback for using BLEUVar is its time complexity, as it needs to calculate the BLEU score between $O(N^2)$ pairs of sequences given $N$ MC samples.

\section{Future Directions}
\label{sect:concs}

With the new tools above we can now develop NMT systems which can be deployed in scenarios where high trust is required of the system, for example in legal applications. With the new tools proposed we can integrate expert annotation in the deployment phase of the system by referring uncertain sentences to human annotation instead of automatic one. Further, with these new tools future research could examine human-in-the-loop approaches to NLP. Such approaches will allow us to develop NLP tools in scenarios when hand-labelling of data is too costly, for example language with scarce resources.

% \begin{itemize}
    % \item From the results of BS in in-distribution experiments, we know that Transformers are well calibrated.
    % \item From the results of BLEUVar in out-of-distribution experiments, we know that BS failed to capture the uncertainty of the model, and BLEUVar is a much more stable uncertainty estimate for both in-distribution and OOD test data.
    % \item Drawback: Number of forward passes: $O(n)$ as it requires $n$ samples during MC dropout. Time complexity for calculating BLEUVar: $O(n^2)$ pairs of sequences given $n$ samples. 
% \end{itemize}


\newpage

\section*{Broader Impact}
% In the paper, we introduce a method to estimate the epistemic uncertainty of NMT models. Our BLEUVar provides an effective uncertainty measure for NMT models to detect out-of-distribution input. Such 

% a) who may benefit from this research
Both the industry and the users of NMT applications can potentially be benefit from our purposed methods. For the industry, an effective uncertainty estimate can be used as a criterion to select the most informative data to be labelled for active learning, where a NMT model can be trained efficiently with a limited amount of labelled data. This is important for the NMT industry, where it is difficult and expensive to acquire labelled training data as human translators are involved in labelling the data. 

\begin{figure}[h]
    \centering
    \includegraphics[width=0.7\textwidth]{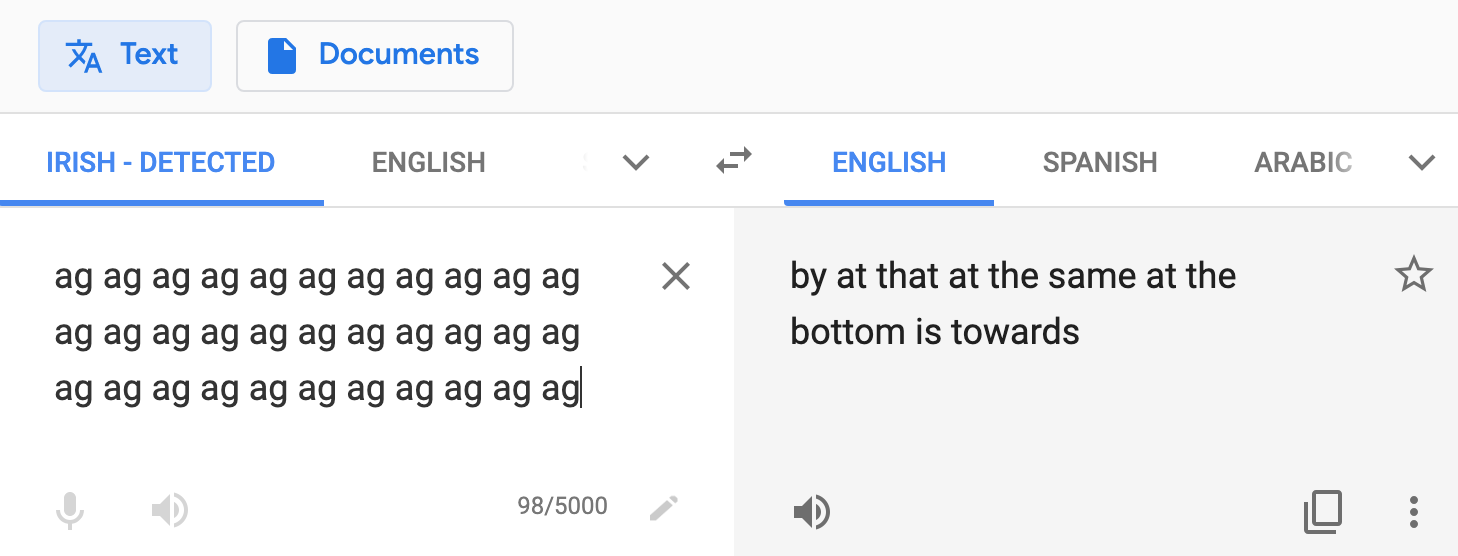}
    \caption{A screenshot from Google Translate for OOD data.}
    \label{bi:google}
\end{figure}

Moreover, the ability to detect the uncertain translations is important for the NMT users. In many scenarios where a large amount of texts need to be translated (e.g. legal documents, literature), people still prefer human translator to machine translation. One reason is that even though the current NMT models have achieved amazing performance for in-distribution data, they are not able to express their uncertainties when a given input is out-of-distribution. For these uncertain input (OOD data), the model is unlikely to generate a good translation (e.g. Figure \ref{bi:google}). Given a high uncertainty score, a NMT model can reject a translation and notified the user that it is not certain about an translation. Our BLEUVar provides an effective uncertainty measure for NMT models to detect out-of-distribution input. BLEUVar is very likely to help the industry and the users of NMT applications to improve their experience in developing and using the latest state-of-the-art NMT models. 

% b) who may be put at disadvantage from this research

% c) what are the consequences of failure of the system
Although our method is designed to identify translation failures in NMT systems, one limitation is that it might underestimate the model uncertainty, which is known for all variational inference methods \cite{barber2011bayesian}. Therefore, it is possible that some out-of-distribution input are not captured by our uncertainty estimate. Fortunately, NMT models are not usually used in mission critical system. Nevertheless, more work is needed in investigating NMT uncertainty if we want to make the most out of the powerful NMT models.

% d) whether the task/method leverages biases in the data
As for bias: A biased model, in many cases, is caused by biases in the training data. It implies that from the perspective of the biased model, the unbiased data is out-of-distribution. Since our purposed uncertainty estimate can detect out-of-distribution data, it might have the potential to be used in investigating whether the trained model is biased or not. Therefore, our method might be able to help reduce bias in NMT model.

% Authors are required to include a statement of the broader impact of their work, including its ethical aspects and future societal consequences. 
% Authors should discuss both positive and negative outcomes, if any. For instance, authors should discuss a) 
% who may benefit from this research, b) who may be put at disadvantage from this research, c) what are the consequences of failure of the system, and d) whether the task/method leverages
% biases in the data. If authors believe this is not applicable to them, authors can simply state this.

% Use unnumbered first level headings for this section, which should go at the end of the paper. {\bf Note that this section does not count towards the eight pages of content that are allowed.}

\bibliography{iclr2020_conference}
\bibliographystyle{abbrvnat}

\newpage

\appendix
\section{In-Distribution Experiments}
\label{sect:in-dist}

This section details our experiments on data that lays within the training distribution for the WMT English (EN) \(\rightarrow\) German (DE) and English (EN) \(\rightarrow\) Vietnamese (VI) tasks. We explore the calibration of Transformer models in this setting and evaluate the effectiveness of using MC Dropout %to improve calibration.
and the proposed methods to measure the model uncertainty. 

In addition to WMT13 dataset for EN $\rightarrow$ DE tasks mentioned in the previous section, we use the \textbf{IWSLT 2015} dataset for translation tasks from EN to VI. There are $133k$ sentences pairs in the \textbf{IWSLT 2015} training set and $1.3k$ sentences pairs in the \textbf{IWSLT 2015} test set. Both the training and test data for \textbf{IWSLT 2015} come from the domain of TED talks.

\subsection{Evaluating Model Calibration}

\begin{figure*}[ht!]
    \centering
%% \vspace{-8mm}
    \begin{subfigure}[b]{0.45\textwidth}
        \centering
        \includegraphics[width=\textwidth]{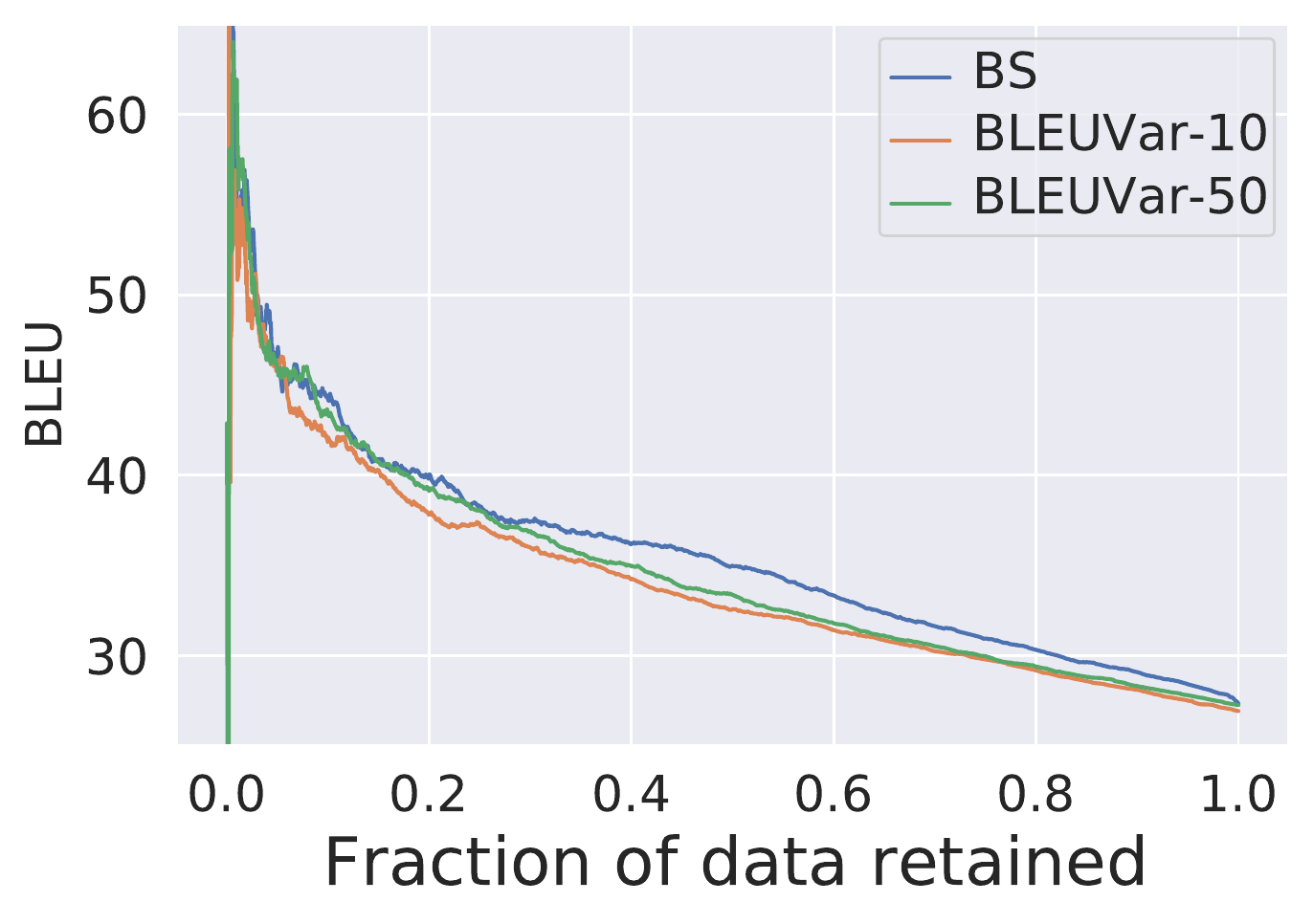}
        \caption{BLEUVar-10 vs BLEUVar-50}
    \end{subfigure}
    \begin{subfigure}[b]{0.45\textwidth}
        \centering
        \includegraphics[width=\textwidth]{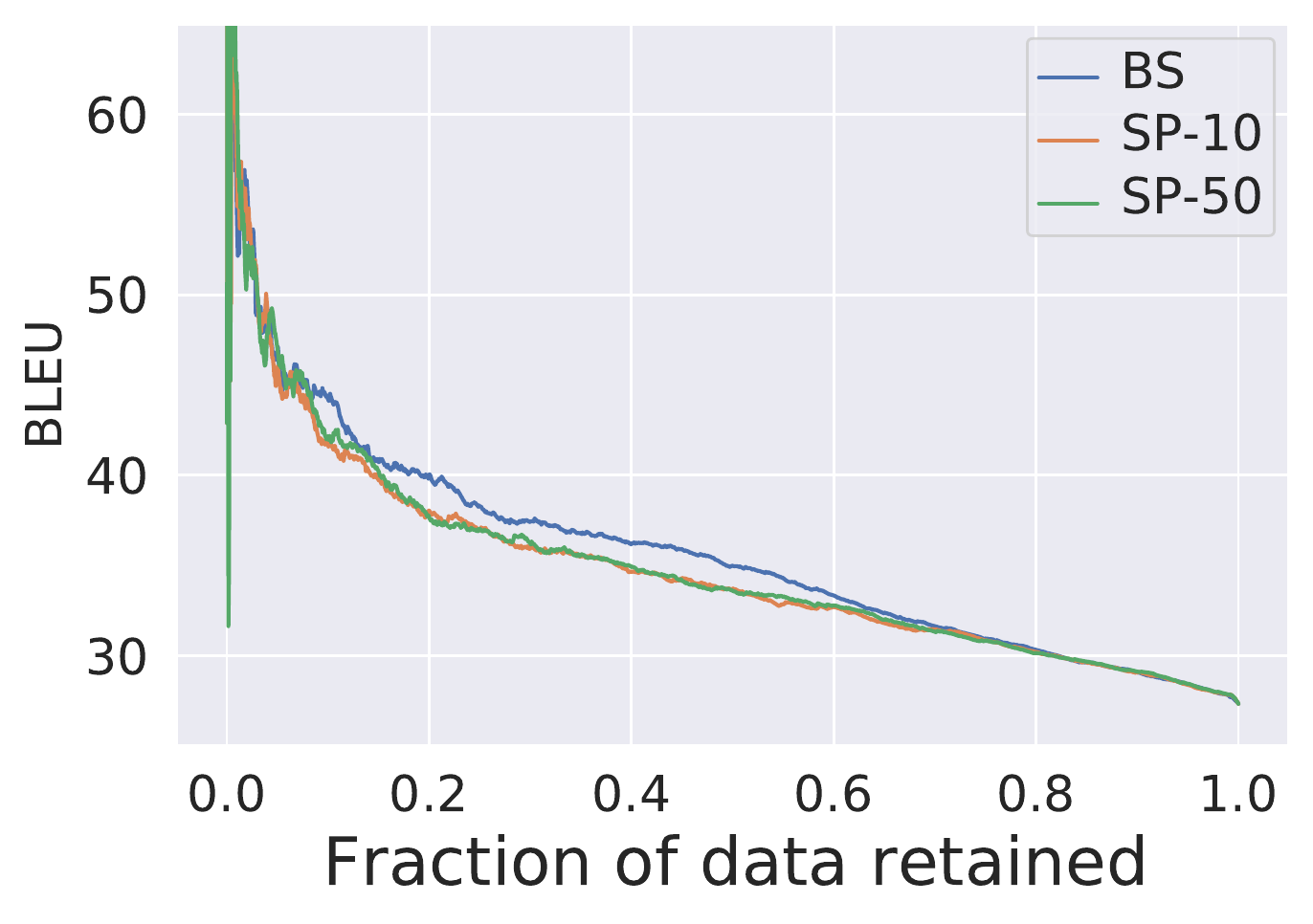}
        \caption{SP-10 vs SP-50}
    \end{subfigure}
    \caption{Uncertainty estimator comparisons for different number of samples. The model is trained for EN to DE tasks with $4.6m$ training data using $350k$ steps.}
    \label{fig:ende-vs50}
\end{figure*}

The first question we hope to answer is the quality of calibration in Transformers models and to evaluate the effectiveness of MC Dropout in improving uncertainty estimates. 

The Transformer was trained on the full EN-DE training set (4.6 million samples) for $350k$ steps. We evaluate on the \textit{newstest2014} test set.

\begin{figure}[ht!]
    \centering
%% \vspace{-8mm}
    \includegraphics[width=0.45\textwidth]{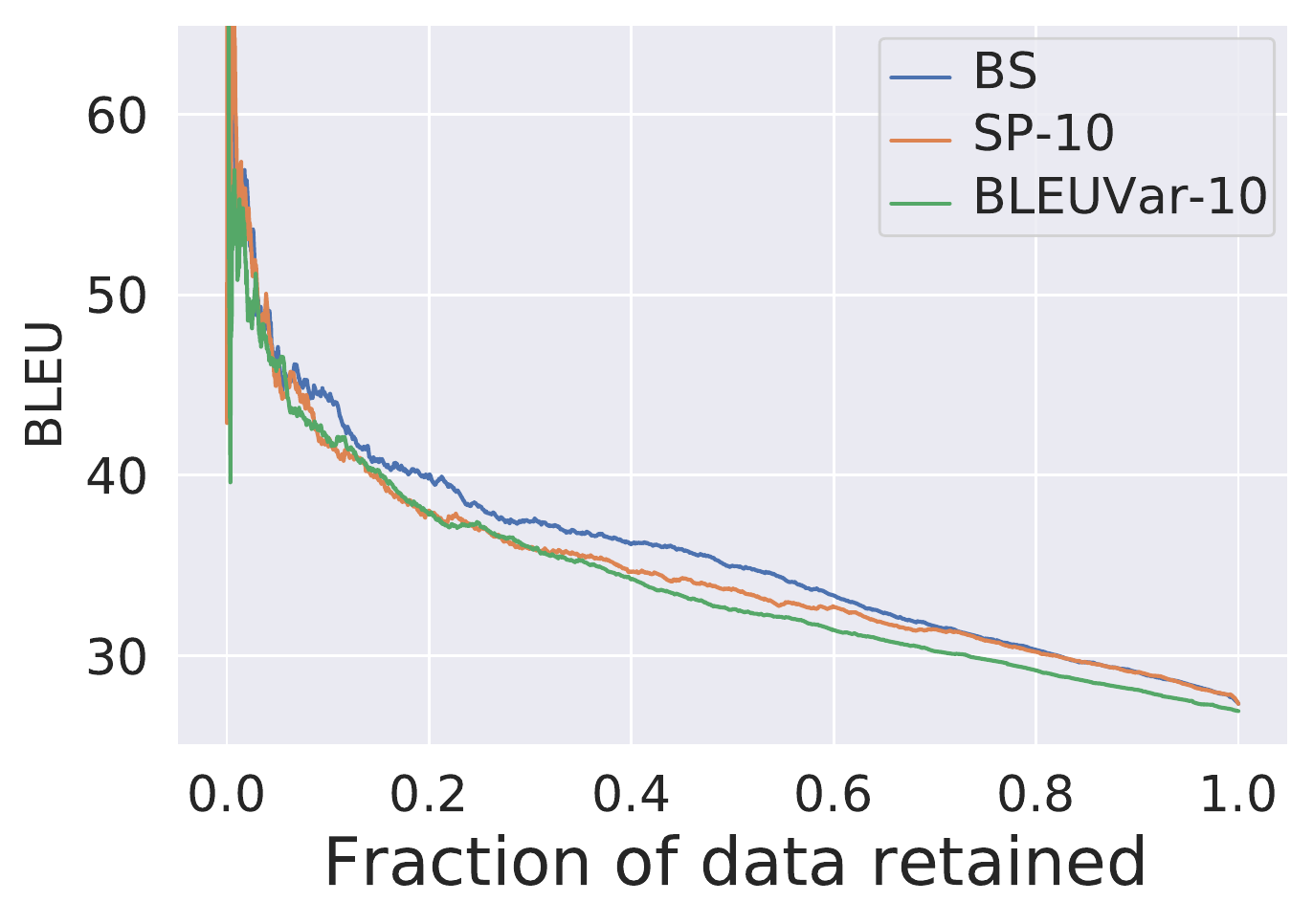}
    \caption{BLEU scores for different uncertainty estimators under various retained data rates. The model is trained for EN to DE tasks with $4.6m$ training data using $350k$ steps.}
    \label{fig:ende-350k-4m-mc10}
\end{figure}

The results from Figures \ref{fig:ende-vs50} and  \ref{fig:ende-350k-4m-mc10} suggest that the beam search score provides a well-calibrated uncertainty metric on the in-distribution test data. The second observation is that MC Dropout-based methods seem to slightly under-perform beam score in this setting, even when the number of samples is increased fivefold. In this setting, our proposed metric (BLEUVar) benefits more from increasing the number of dropout samples relative to sequence probability.

\subsection{The Impact of Training Set Size}
\label{sect:train-size}

\begin{figure*}[ht!]
%% \vspace{-8mm}
    \centering
    \begin{subfigure}[b]{0.45\textwidth}
        \centering
        \includegraphics[width=\textwidth]{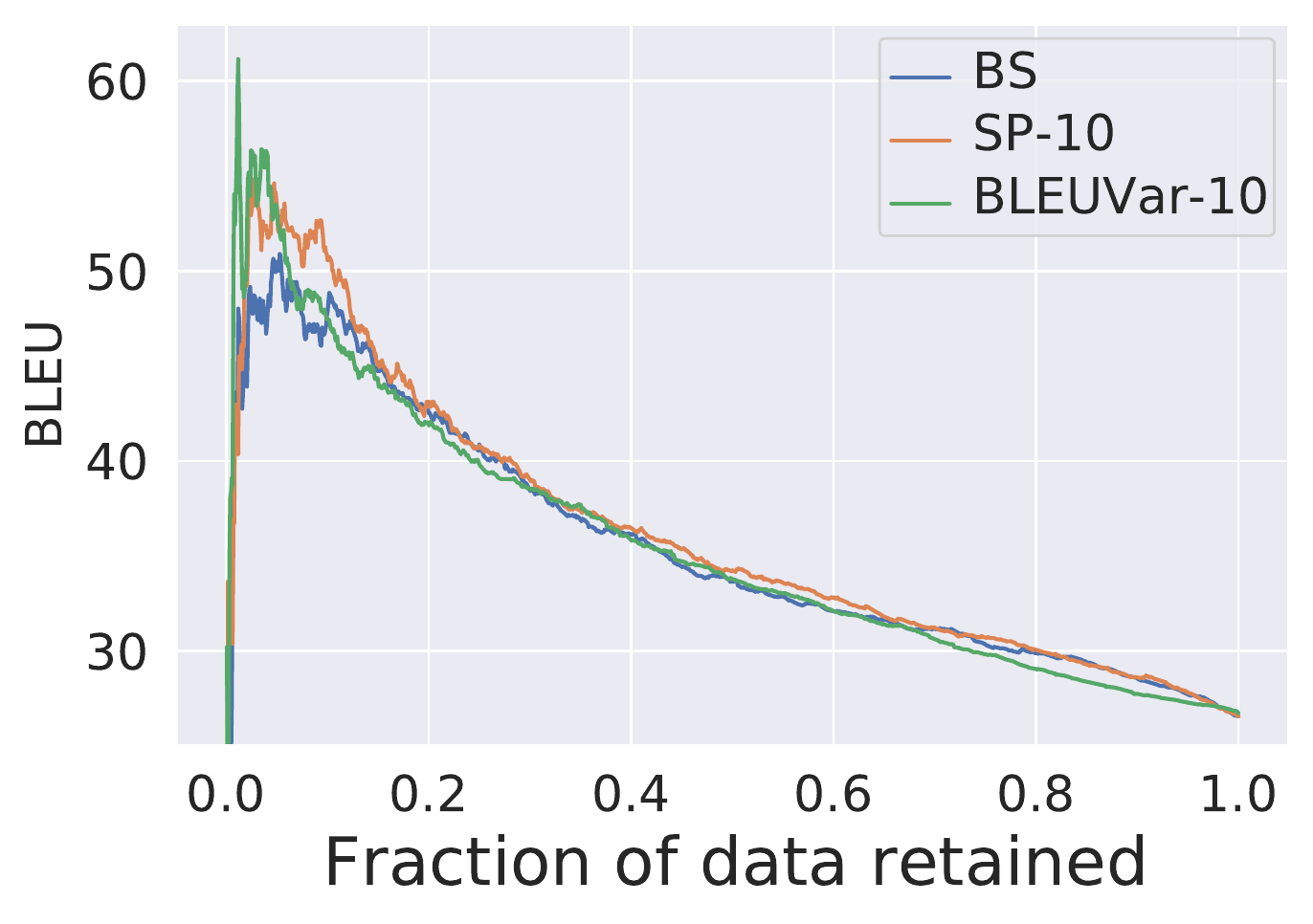}
        \caption{10 samples}
    \end{subfigure}
    \begin{subfigure}[b]{0.45\textwidth}
        \centering
        \includegraphics[width=\textwidth]{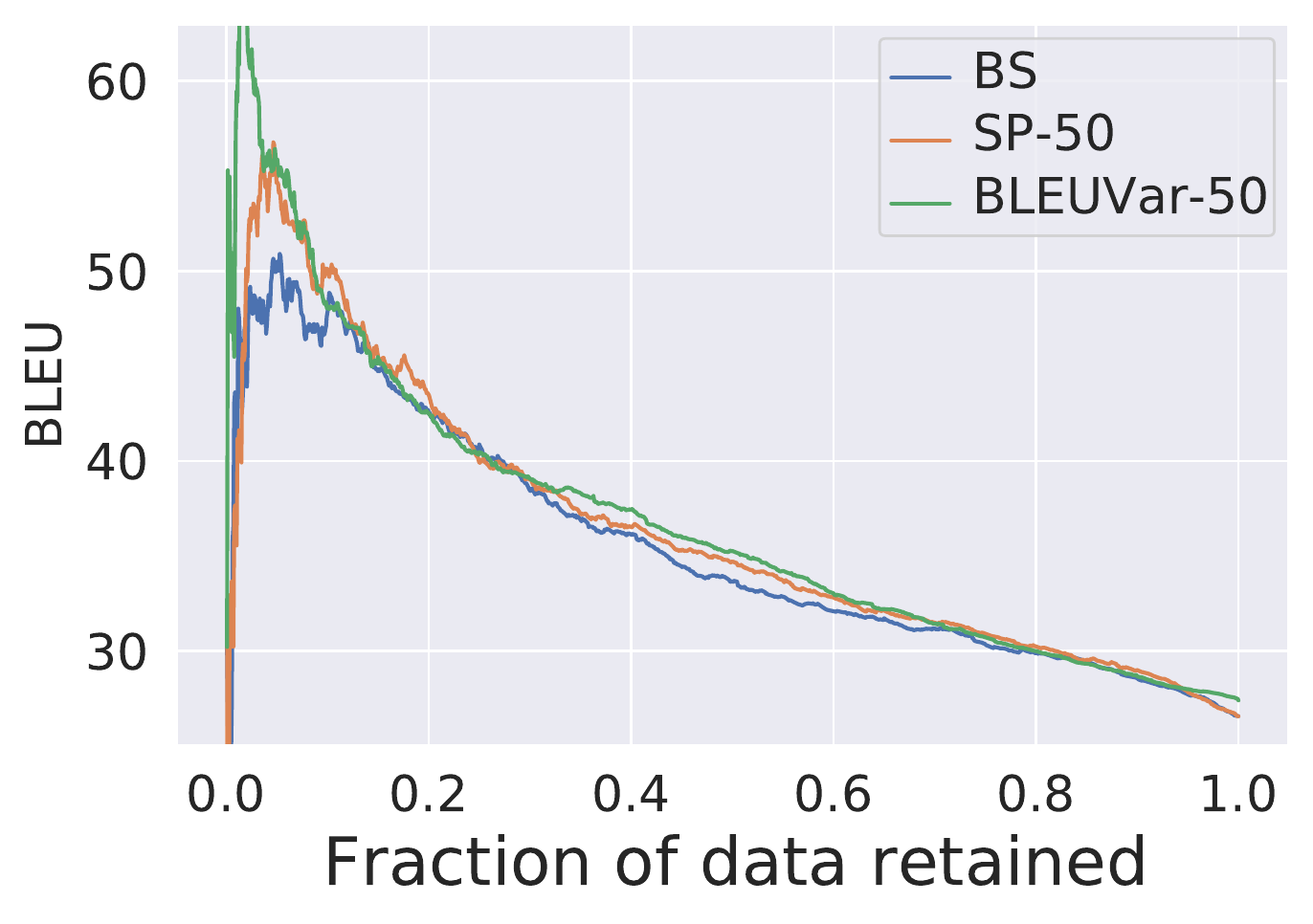}
        \caption{50 samples}
    \end{subfigure}
    \caption{BLEU scores for different uncertainty estimators under various retained data rates. The model is trained for EN to VI tasks with $133k$ training data using $350k$ steps.}
    \label{fig:envi-350k-mc10-50}
\end{figure*}

The WMT EN-DE training set is fairly large and one would assume that most test sentences (or very similar ones) have been observed during training time. Hence we do not expect much epistemic uncertainty to exist in this testing scenario, which the experiments seem to confirm.
A natural question to ask is on the effect of training set size on the calibration of models. We explore this question by considering the WMT English to Vietnamese (EN-VI) task which has $133k$ samples in the training set (approx. 2.6\% of EN-DE), and down-sampling the EN-DE training set to $50k$ and $100k$ samples.

The performance-retention plots in Figure \ref{fig:envi-350k-mc10-50} indicate that, while a large training set yields curves that seem to suggest beam score is a sufficient uncertainty metric, when a small dataset is used the MC Dropout-based uncertainty metrics begin to outperform the beam score (note the retention range 0.0 to 0.2). Moreover, in the low data setting increasing the number of samples drawn from MC dropout results in a significant improvement for BLEUVar.

% Take away:
% \begin{itemize}
%     \item The three methods can work in different language pair settings.
%     \item Both SP and BLEUVar can outperform BS.
%     \item Again, increasing number of samples can significantly improve the performance of BLEUVar.
%     \item The three methods have similar performance for in-distribution test sets.
% \end{itemize}

\begin{figure*}[ht!]
% \vspace{-3mm}
    \centering
    \begin{subfigure}[b]{0.45\textwidth}
        \centering
        \includegraphics[width=\textwidth]{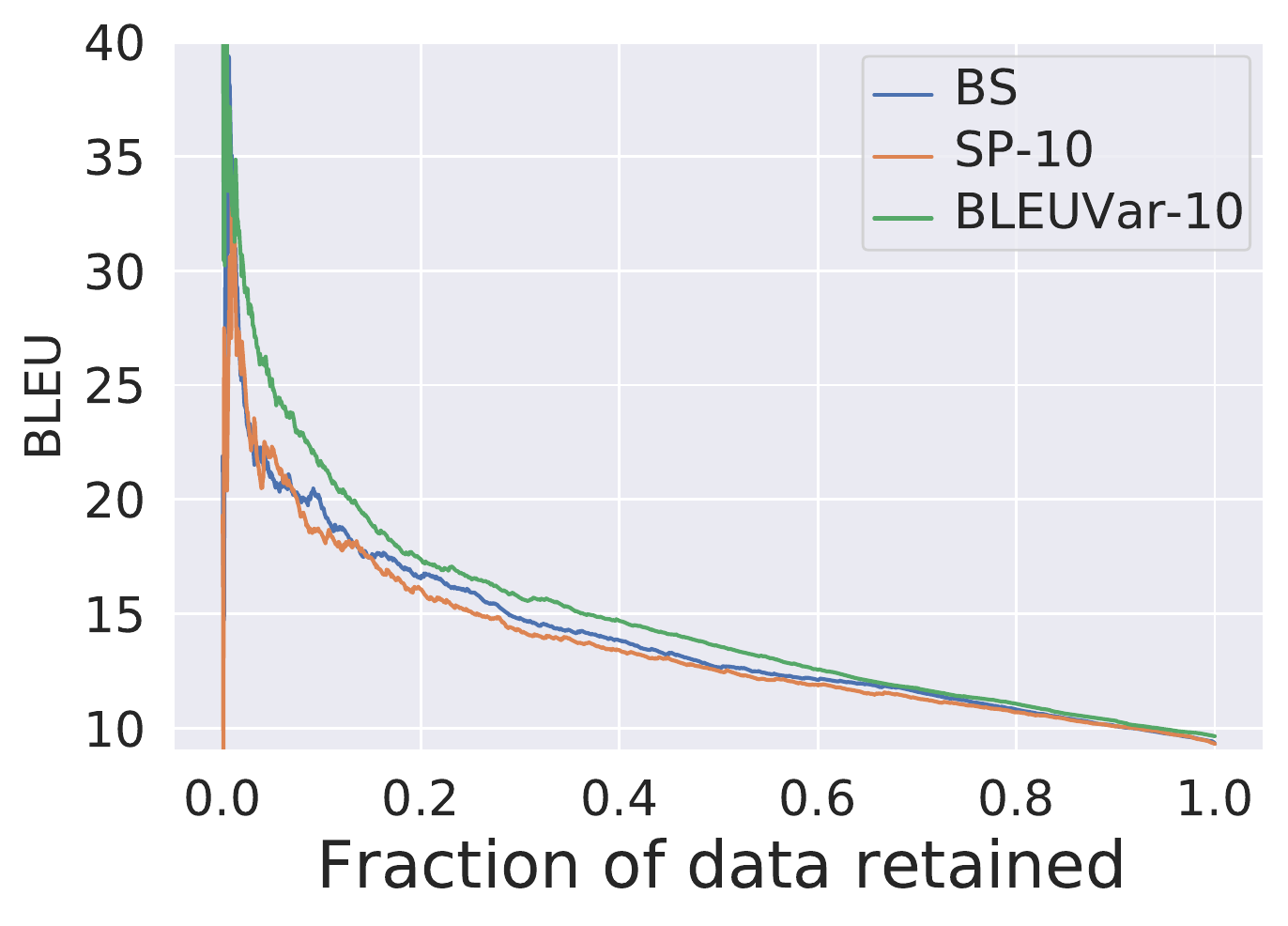}
        \caption{$50k$ training data}
    \end{subfigure}
    \begin{subfigure}[b]{0.45\textwidth}
        \centering
        \includegraphics[width=\textwidth]{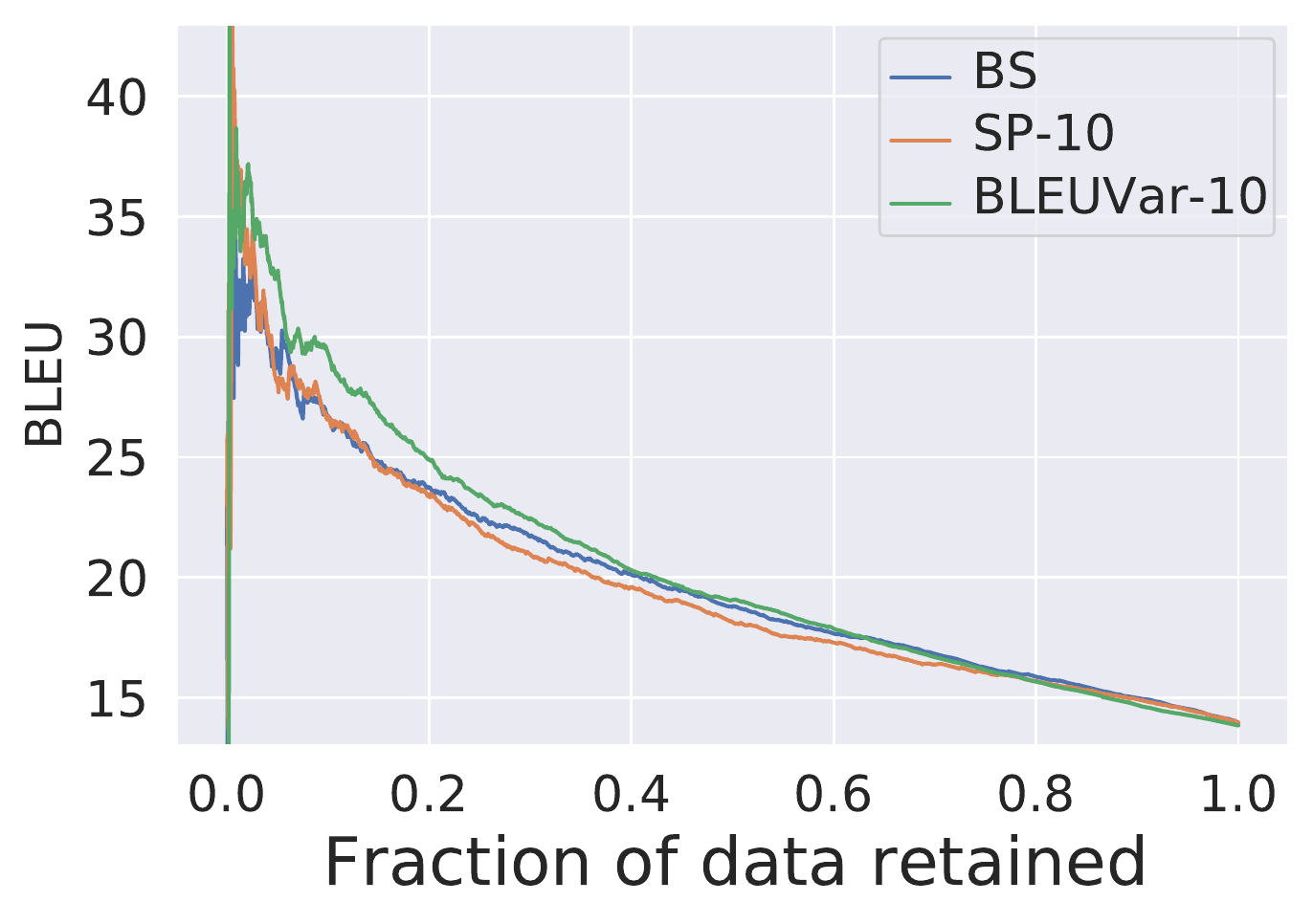}
        \caption{$100k$ training data}
    \end{subfigure}
    \caption{Uncertainty estimator comparisons for models with different sizes of training set. The models were trained for EN to DE tasks with $50k$ and $100k$ training data using $350k$ steps.}
    \label{fig:ende-t50-100k}
\end{figure*}

% \begin{figure}[ht!]
%     \centering
%     \includegraphics[width=0.3\textwidth]{sb-ende-350k-50k-mc10-individual-scatter.png}
%     \caption{The density of individual BLEU score versus uncertainty for all sentences. The sentences are ordered by their uncertainty from low (left) to high (right) using BLEUVar-10. Following the calculation of BLEUVar, since we have 10 samples, the uncertainty BLEUVar-10 has the value in range $[0,90]$. And we multiply them with 100, which results in the x-axis has the range $[0,9000]$. The model was trained for EN to DE tasks with $50k$ training data using 350k steps.}
%     \label{fig:ende-350k-50k-individual}
% \end{figure}

\begin{figure*}[ht!]
% %\vspace{-8mm}
    \centering
    \begin{subfigure}[b]{0.47\textwidth}
        \centering
        \includegraphics[width=\textwidth]{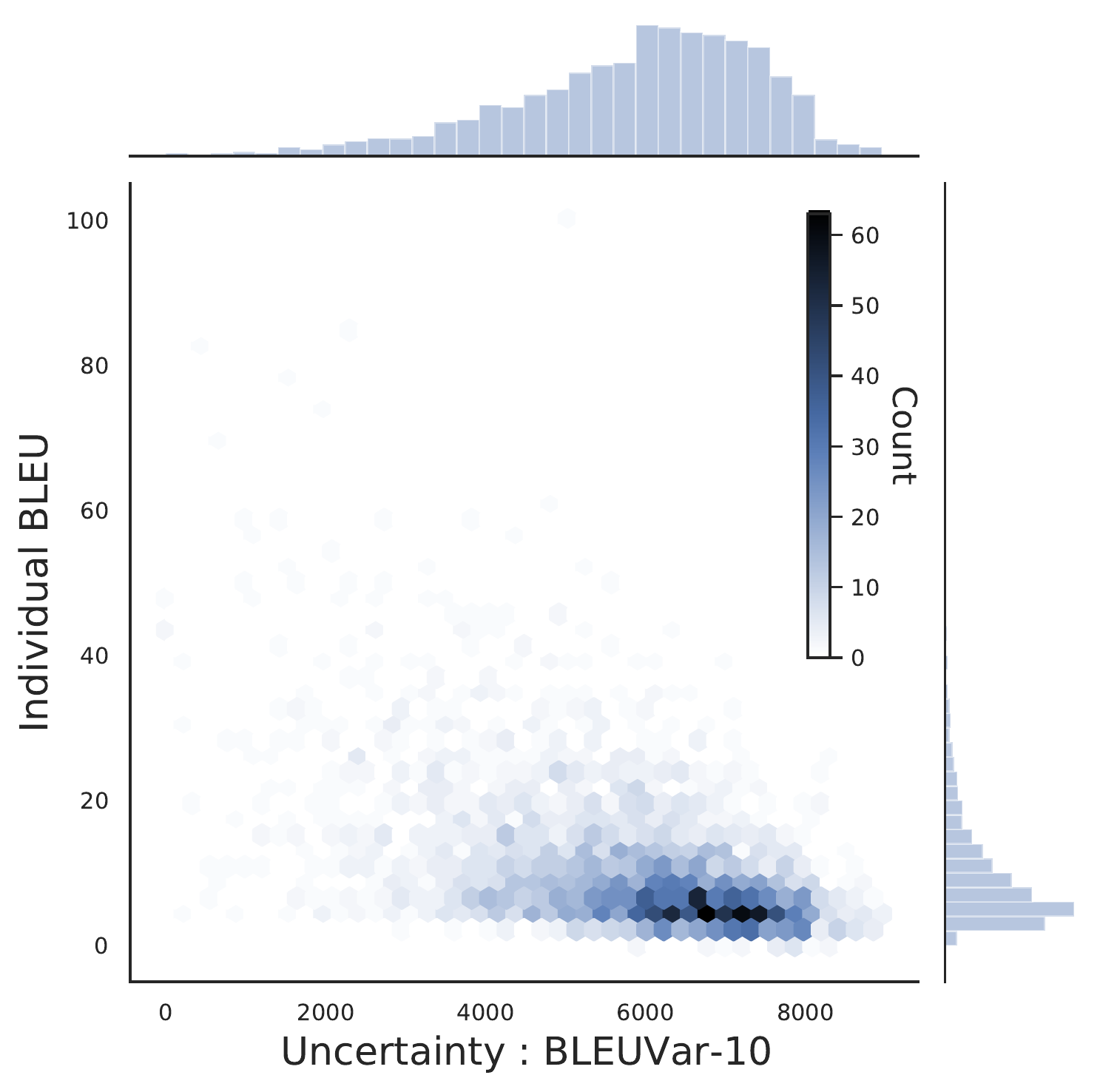}
        \caption{$50k$ training data}
    \end{subfigure}
    \begin{subfigure}[b]{0.47\textwidth}
        \centering
        \includegraphics[width=\textwidth]{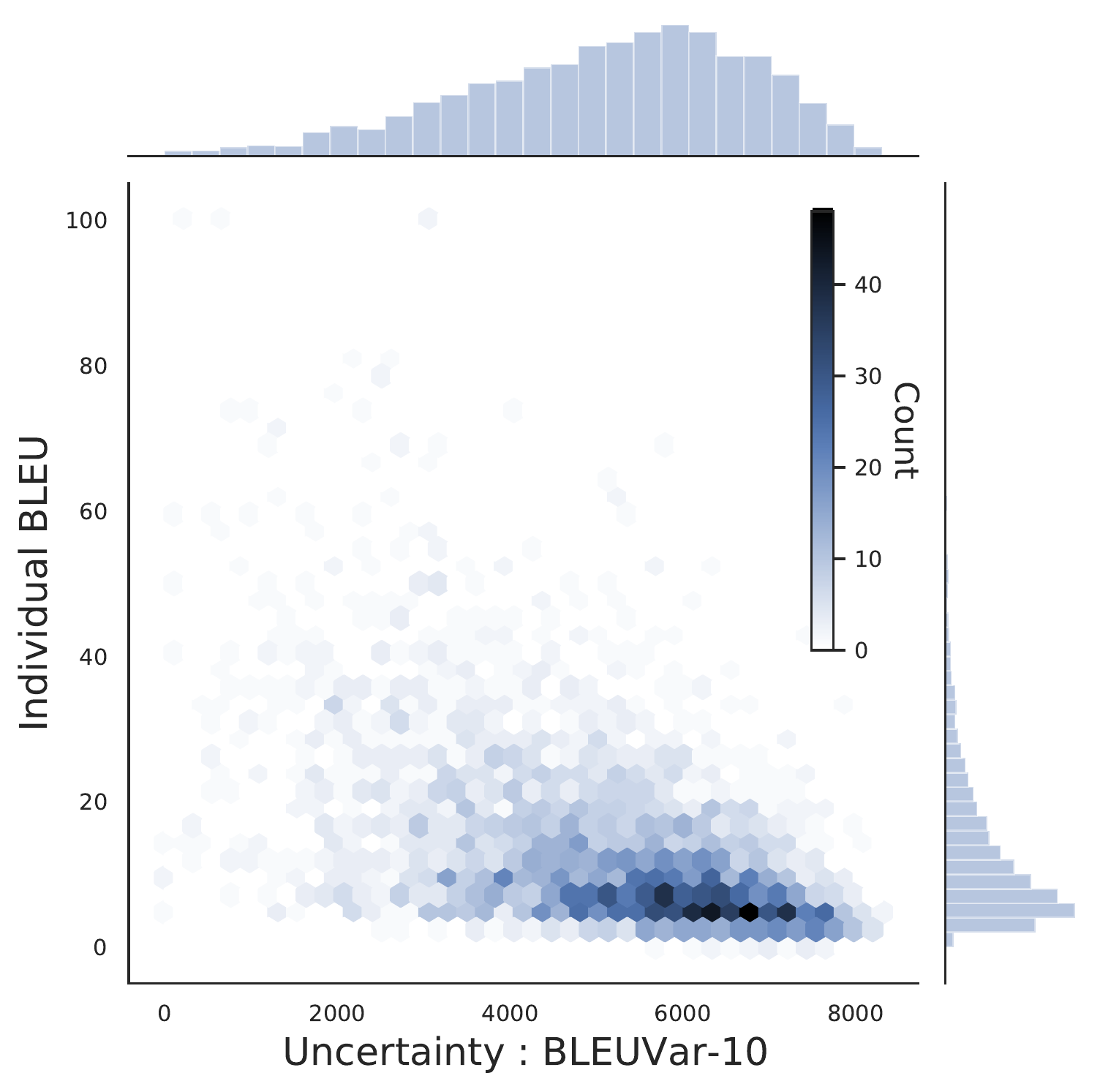}
        \caption{$100k$ training data}
    \end{subfigure}
    \begin{subfigure}[b]{0.47\textwidth}
        \centering
        \includegraphics[width=\textwidth]{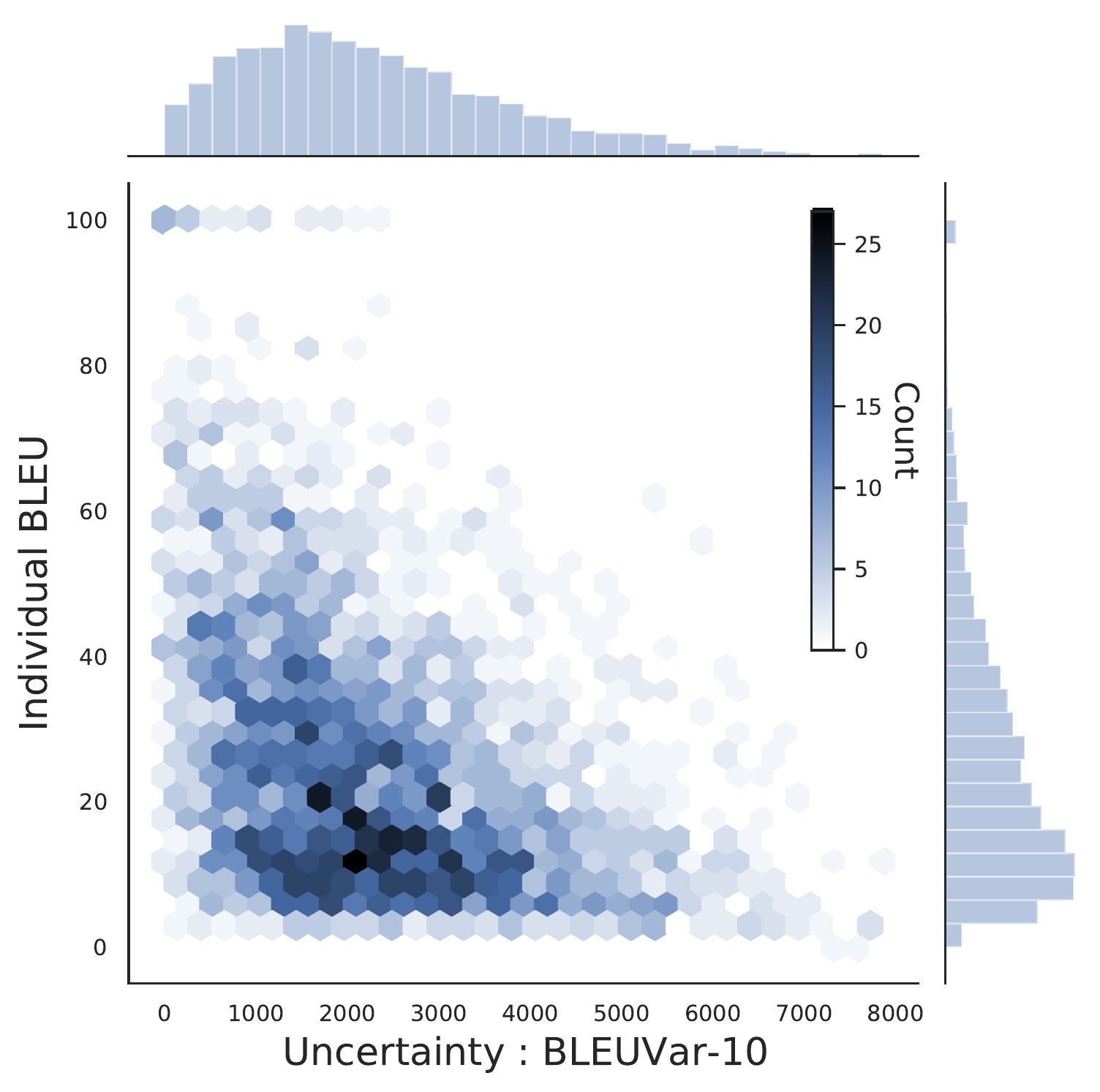}
        \caption{$4.6m$ training data}
    \end{subfigure}
    \caption{
    The density of individual BLEU score versus uncertainty (BLEUVar-10) for all sentences in the same test set \emph{newstest2014} produced by models trained with various size of data set. The sentences are ordered by their uncertainty from low (left) to high (right) using BLEUVar-10. Following the calculation of BLEUVar, since we have 10 samples, the uncertainty estimate BLEUVar-10 has the value in range $[0,90]$. And we scale it up by $\times 100$, which results in the x-axis has the range $[0,9000]$. The models were trained for EN to DE tasks with $50k$, $100k$ and $4.6m$ training data using $350k$ steps.}
    \label{fig:joints}
\end{figure*}

The experiments depicted in Figures \ref{fig:ende-t50-100k} and \ref{fig:joints} consist of down-sampling the EN-DE training set. 
Figure \ref{fig:ende-t50-100k} demonstrates a similar pattern to the above EN-VI experiment when down-sampling the EN-DE data to $50k$ and $100k$ examples. Again, in the low-data regime BLEUVar substantially out-performs beam score and sequence probability.
Figure \ref{fig:joints} demonstrates the impact of data size on the distribution of example uncertainty versus performance. We see that low data regimes lead to a low-entropy distribution with high uncertainty across the entire test set; as data availability is increased, uncertainty decreases, and average model performance increases for all rates of data retention.
% and a more reasonable lower-triangular density forms.

% Take away:
% \begin{itemize}
%     \item With smaller training set \emph{BLEU variance} outperforms \emph{beam score} and \emph{sequence probability}.
%     \item Figure \ref{fig:joints}: with more training data, the model is more confident about the same input. 
%     \item Figure \ref{fig:joints}: The bottom left of each subplot corresponds to high confidence but with low bleu scores, which means if a sentence can be translated with high BLEU score, it will be assigned with high confidence, but if a sentence is assigned with high confidence, it doesn't imply that it will have a high BLEU score. A possible explaination is that one sentence from language A can be translated into many different sentences in language B with the same meaning, which are hard to capture using BLEU. With higher confidence, the model is more likely to get the `perfect' answer (higher BLEU), but not necessarily (therefore the density in the lower half bottom right triangle). On the other hand, with higher uncertainty, it's much less possible to get the `perfect' answer. It can also be a matter of underestimating uncertainty for some sentences.
% \end{itemize}

\section{Related Work}
\label{sect:related}

Our work might look similar to quality estimation (QE) task in MT \citep{specia2010machine, blatz2004confidence}, but the problem of QE is fairly different to what we do in this paper. QE assumes the existence of a fixed translation system (e.g., an in-house encoder-decoder attention-based NMT system, as in WMT19’s shared task in Quality Estimation). The QE models then have to determine the quality of the system’s output. In contrast, we look at the problem of “introspection” where the system has to decide the the confidence (“quality”) of \textit{its own} output. This confidence can then be used for selective classification where the model can reject some uncertain translation. Further, standard approaches in QE might assume access to privileged data (e.g., the NMT translations for the source sentences and their corresponding human post-edition, as in task 2 in WMT19’QE), which we do not require. In addition, most existing approaches for QE require additional model to be trained to estimate the translation quality of a MT model, while our method does not have such requirement. Therefore, our method is able to provide uncertainty estimate simply with the parallel corpus used for training the translation model without the need for additional data and training procedure.

The closest to our paper is task 3 in WMT19’QE: a metric to score sentences is sought, which must correlate to human judgement. We would like to stress that a system’s confidence in its own prediction does not have to be correlated to human judgement. Indeed, we demonstrate this in Appendix \ref{sect:train-size} where a model can indicate that it does not have enough training data, and requires additional data to increase its confidence (the model’s subjective view of its uncertainty does not have to correlate with empirical mistakes - Bayesian epistemology \citep{zalta1995stanford}). 

% \tim{Need review here. Work in progress waiting for the QE result}
% The setting for QE is quite different from our task. The training data provided by WMT QE consists of source sentences, the NMT translations and the post-edited sentences for the translations. In contrast, the Transformer model we used requires parallel corpus (i.e. set of source sentences and target sentences) therefore the QE training set cannot be used to train our model 

In addition, QE tasks are mainly focus on estimating the in-distribution translation quality, since the test sets are in the same domain as the training sets provided by WMT QE tasks (e.g. both in the IT domain for English-German WMT18,19). In contrast, the goal for our uncertainty estimate is to identify the out-of-distribution translations, rather then estimating the quality of in-distribution translation. Therefore, our tasks are fundamentally different to QE.
% We did an experiment to test a QE method in the out-of-distribution setting. The QE model we used is the predictor-estimator \citep{kim2017predictor}, which is the QE baseline system recommended by WMT20' for sentence-level quality estimation tasks. We used 4 pre-trained predictor-estimator models from OpenKiwi \citep{openkiwi}, which were trained on English-German sentences from the IT domain. \tim{Work in progress, it depends on whether we do a new experiments for comparison.}

% \begin{figure*}[ht!]
% % \vspace{-4mm}
%     \centering
%     \includegraphics[width=0.7\textwidth]{qe-ende-350k-mc10.png}
%     \caption{\tim{This might need to be changed!} Our uncertainty measures versus 4 different QE Predictor-Estimator models \citep{kim2017predictor} using the test set \emph{newstest2014}. The Transformer model was trained for EN to DE tasks with the WMT13' dataset using $350k$ steps, the Predictor-Estimator models were trained in ... .}
%     \label{fig:qe}
% \end{figure*}

There have been some prior attempts at investigating the similar problem as ours. In particular, \citet{kumar2019calibration} investigated the calibration of various NMT models at the token level. \citeauthor{kumar2019calibration} found that most models are ill-calibrated at the token level, leading to the resulting probability distribution over the vocabulary used during decoding is not a good reference for model uncertainty. To correct for this, \citeauthor{kumar2019calibration} design a recalibration strategy that applies an adaptive temperature to the logits, determined by the token identities, attention entropies, and other relevant components. Another study of uncertainty in NMT models comes from \citet{ott2018scaling}; they found models tend to have overly high uncertainty in their output distribution over sequences.  Note that both do not consider \textit{epistemic} uncertainty, not OOD settings.
Our work explores this direction and offers a new uncertainty estimation technique that empirically out-performs existing methods by a significant margin.

\section{Translation Samples}
\label{sect:samples}

\subsection{In-distribution Certain Samples}
\label{sect:samples:in-dist-de-1}
    \begin{longtable}{l|l|p{5cm}}
      \toprule % <-- Toprule here
      \multicolumn{3}{l}{\textbf{Source sentence} (DE) : } \\[0.3em]
      \multicolumn{3}{l}{Nevada hat bereits ein Pilotprojekt abgeschlossen.}\\
      \midrule % <-- Midrule here
      \multicolumn{3}{l}{\textbf{Reference translation} (EN) : {\small(only used to compute “BLEU to reference”)
}} \\[0.3em]
      \multicolumn{3}{l}{Nevada has already completed a pilot.}\\
      \midrule % <-- Midrule here
      \multicolumn{3}{l}{\textbf{Model predictive-mean translation} (EN) : {\small(averaging over predictive probabilities during decoding)
} } \\[0.3em] 
      \multicolumn{3}{l}{Nevada has already completed a pilot project.}\\
      \midrule % <-- Midrule here
      \multicolumn{2}{l|}{\textbf{Translation “BLEU to reference”} : } & 70.7\\
      \midrule % <-- Midrule here
      \multicolumn{2}{l|}{\textbf{Translation uncertainty} : } & 0\\
      \midrule % <-- Midrule here
      
      \multicolumn{3}{l}{\textbf{Translations sampled from the model}: {\small(5 samples from predictive probabilities during decoding)}} \\
      \midrule % <-- Midrule here
      \textbf{1} & \multicolumn{2}{l}{Nevada has already completed a pilot project.} \\
      \textbf{2} & \multicolumn{2}{l}{Nevada has already completed a pilot project.} \\
      \textbf{3} & \multicolumn{2}{l}{Nevada has already completed a pilot project.} \\
      \textbf{4} & \multicolumn{2}{l}{Nevada has already completed a pilot project.} \\
      \textbf{5} & \multicolumn{2}{l}{Nevada has already completed a pilot project.} \\

      \bottomrule % <-- Bottomrule here
    % \end{tabular}
    \caption{\Tstrut (Low uncertainty) In-distribution DE source sentence  from the experiment in Figure \ref{fig:de-en-nl-en}(a).}
    \label{tab:in-dist-de-1}
    \end{longtable}
%   \end{center}
% \end{table}

\newpage
\subsection{In-distribution Uncertain Samples}
\label{sect:samples:in-dist-de-2}

% \begin{table}[ht!]
%   \begin{center}
    \begin{longtable}{l|l|p{5cm}}
      \toprule % <-- Toprule here
      \multicolumn{3}{l}{\textbf{Source sentence} (DE) : } \\[0.3em] 
      \multicolumn{3}{l}{Im Grunde genommen sind vegane Gerichte für alle da.}\\
      \midrule % <-- Midrule here
      \multicolumn{3}{l}{\textbf{Reference translation} (EN) : {\small(only used to compute “BLEU to reference”)
}} \\[0.3em]
      \multicolumn{3}{l}{Essentially, vegan dishes are for everyone.}\\
      \midrule % <-- Midrule here
      \multicolumn{3}{l}{\textbf{Model predictive-mean translation} (EN) : {\small(averaging over predictive probabilities during decoding)
}} \\[0.3em]
      \multicolumn{3}{l}{Basically vegan dishes are there for everyone.}\\
      \midrule % <-- Midrule here
      \multicolumn{2}{l|}{\textbf{Translation “BLEU to reference”} : } & 34.5\\
      \midrule % <-- Midrule here
      \multicolumn{2}{l|}{\textbf{Translation uncertainty} : } & 3122\\
      \midrule % <-- Midrule here
      
      \multicolumn{3}{l}{\textbf{Translations sampled from the model}: {\small(5 samples from predictive probabilities during decoding)}} \\
      \midrule % <-- Midrule here
      \textbf{1} & \multicolumn{2}{l}{Basically vegan dishes are for everyone.} \\
      \textbf{2} & \multicolumn{2}{l}{Basically, vegan dishes are there for everyone.} \\
      \textbf{3} & \multicolumn{2}{l}{Essentially, vegan dishes are available for everyone.} \\
      \textbf{4} & \multicolumn{2}{l}{Basically, vegane dishes are there for all.} \\
      \textbf{5} & \multicolumn{2}{l}{Basically vegan dishes are there for everyone.} \\

      \bottomrule % <-- Bottomrule here
    % \end{tabular}
    \caption{\Tstrut (High uncertainty) In-distribution DE source sentence from the experiment in Figure \ref{fig:de-en-nl-en}(a).}
    \label{tab:in-dist-de-2}
    \end{longtable}
%   \end{center}
% \end{table}

% \newpage
\subsection{Out-of-distribution Samples}
\label{sect:samples:ood-nl}

% \begin{table}[ht!]
%   \begin{center}
    \begin{longtable}{l|l|p{6cm}}
      \toprule % <-- Toprule here
      \multicolumn{3}{l}{\textbf{Source sentence} (NL) : } \\[0.3em]
      \multicolumn{3}{p{13cm}}{De debiteurenlanden zouden hun concurrentiekracht terugkrijgen; hun schulden zouden in reële termen afnemen; de dreiging van staatsbankroeten zou - met de ECB onder hun controle - verdwijnen, en hun leenkosten zouden dalen naar een niveau dat vergelijkbaar is met dat van het Verenigd Koninkrijk.}\\
      \midrule % <-- Midrule here
      \multicolumn{3}{l}{\textbf{Reference translation} (EN) : {\small(only used to compute “BLEU to reference”)
}} \\[0.3em]
      \multicolumn{3}{p{13cm}}{Debtor countries would regain their competitiveness; their debt would diminish in real terms; and, with the ECB under their control, the threat of default would disappear and their borrowing costs would fall to levels comparable to that in the United Kingdom.}\\
      \midrule % <-- Midrule here
      \multicolumn{3}{l}{\textbf{Model predictive-mean translation} (EN) : {\small(averaging over predictive probabilities during decoding)
}} \\[0.3em] 
    \multicolumn{3}{p{13cm}}{The debitenlands were to compete with the rivalrivalrivalrivalrivalrivalrivalrivals of terugkrijgen; they were in debt in the countries of \seqsplit{afafafafafafafafafafafafafafafafafafafafafafafafafafafafafafafafafafafafafafafafafafafafafafafafafafafafafafafafafafafafafafafafafafafafafafafafafafafafafafafafafafafafafafafafafafafafafafafafafafafafafafafafafafafafafafafafafafafafafafafafafafafafafafafafafaf}}\\
      \midrule % <-- Midrule here
      \multicolumn{2}{l|}{\textbf{Translation “BLEU to reference”} : } & 1.9\\
      \midrule % <-- Midrule here
      \multicolumn{2}{l|}{\textbf{Translation uncertainty} : } & 8617\\
      \midrule % <-- Midrule here
      
      \multicolumn{3}{l}{\textbf{Translations sampled from the model}: {\small(5 samples from predictive probabilities during decoding)}} \\
      \midrule % <-- Midrule here
      \textbf{1} & \multicolumn{2}{p{13cm}}{The debitenlands were the ones to compete in their rivalrivalrivalrivalrivalrivalrivalrivalrivalrivalrivalrivals of them; they were debt-denominated in their afafafafafafafafen; the tripthirthirthirwent of state bankrbankrbankrbankrbankrbankrbankrbankrbankrbankrbankrbankrbankrbankrbankru - with the ECB in its control - the run - the run - the run - the run - the run - the run - the run - the run - the run-off - the run - the run-run run run run run of the ECB.} \\
      \textbf{2} & \multicolumn{2}{p{13cm}}{In the \seqsplit{debdebdebdebdebdebdebdebdebdebdebdebdebdebdebdebdebdebdebdebdebdebdebdebdebdebdebdebdebdebdebdebdebits}, the debdebdebdebdebdebdebdebdebits were competitive in terms of law; those debt owed in debt in debt; the three of state \seqsplit{bankrbankrbankrbankrbankrbankrbankrbankrbankrbankrbankrbankrbankrbankrbankrbankrbankrbankrbankrbankrbankrbankrbankrbankrbankrbankrbankrbankrbankrbankrbankrbankrbankrbankrbankrbankrbankrbankrbankrbankrbankrbankrbankr} -  with the ECB, in its, in its, in its, in its control - business - business - the ECB, in its, in the control - the dispute, the - business - the dispute, the dispute, the - business - the dispute, the sovereign} \\
      \textbf{3} & \multicolumn{2}{p{13cm}}{At the time of its independence, it was a rivalrivalrivalrivalrivalrivalrivalrivalrivalrivalrivalrivalrivalrivalrivalrivalrivalrivalrivaleach one; the debts of the poor; the three-three of the bankrbankrbankrall - with the ECB in its control of the ones - the ones in question - the parties in question, the countries in the future; the three of the bankrbankrbankrbankrbankrbankrbankrbankrbankr ( with the ECB in its control, with the ECB in its control, in the face, the disputes, the financial crises, the financial crisis, the financial crisis, the financial crisis, the financial crisis, the financial crisis, the financial crisis, the financial crisis, the financial crisis, the financial crisis, the financial crisis, the financial crisis, the financial crisis.} \\
      \textbf{4} & \multicolumn{2}{p{13cm}}{De debitenlanden zouden hun \seqsplit{concconcurrentierivalrival} terugkrijgen; hun \seqsplit{levlevlevlevlevlevlev} \seqsplit{levlevlevlevlevlevlevlevlevlevlevlevlevlevlevlevlevlevlevlevlevlevlevlevlevlevlevlevlevlevlevlevlevlevlevlevlevlevlevlevlevlevlevlevlevlevlevlevlevlevlevlevlevlevlevlevlevlevlevlevlevlevlevlevlevlevlevlevlevlevlevlevlevlevlevlevlevlevlevlevlevlevlevlevlevlevlevlevlevlevlevlevlevlevlevlevlevlevlevlevlevlevlevlevlevlevlevlevlevlevlevlevlevlevlevlevlevlevlevlevlevlevlevlevlevlevlevlevlevlevlev}} \\
      \textbf{5} & \multicolumn{2}{p{13cm}}{The debitenland \seqsplit{gambgambgambgambgambgambgambgambgambgambgambgambgambgambgambgamble} in their own countries; the \seqsplit{gambgambgambgambgambgambgambgambgambgambgambgambgambgambgambgambgambgambgambgambgambgambgambgambgambgambgambgambgambgambgambgambgambgambgambgambgambgambgambgambgambgambgambgambgambgambgambgambgambgambgambgambgambgambgambgambgambgambgambgambgambgambgambgambgambgambgambgambgambgambgambgambgambgambgambgambgambgambgambgambgambgambgambgambgambgambgambgambgambgambgambgambgambgambgambgambgambgambgambgambgambgambgambgambgambgambgambgambgambgambgambgambgambgambgambgambgambgambgambgambgambgambgambgambgambgambgambgambgambgambgambgambgambgambgambgamb}} \\

      \bottomrule % <-- Bottomrule here
    % \end{tabular}
    \caption{\Tstrut Out-of-distribution NL source sentence from the experiment in Figure \ref{fig:de-en-nl-en}(c). }
    \label{tab:ood-nl}
    \end{longtable}

\end{document}